\documentclass[a4paper,fleqn]{cas-sc}
\pdfoutput=1
\usepackage{subcaption}
\usepackage{graphicx}
\usepackage{adjustbox}
\usepackage{subcaption}
\usepackage{amsmath}
\UseRawInputEncoding
\usepackage{float}  
\usepackage[authoryear]{natbib}
\UseRawInputEncoding
\usepackage[utf8]{inputenc}
\DeclareUnicodeCharacter{202A}{}

\def\tsc#1{\csdef{#1}{\textsc{\lowercase{#1}}\xspace}}
\tsc{WGM}
\tsc{QE}
\tsc{EP}
\tsc{PMS}
\tsc{BEC}
\tsc{DE}

\begin{document}
\let\WriteBookmarks\relax
\def\floatpagepagefraction{1}
\def\textpagefraction{.001}

\shorttitle{Digital Twin Framework for Optimal and Autonomous Decision-Making in Cyber-Physical Systems}

\shortauthors{Rebello, C.M. et~al.}

\title[mode = title]{Digital Twin Framework for Optimal and Autonomous Decision-Making in Cyber-Physical Systems: Enhancing Reliability and Adaptability in the Oil and Gas Industry}




%
\author[1]{Carine Menezes Rebello}[type=editor,
                        orcid=0000-0002-0796-8116]



\ead{carine.m.rebello@ntnu.no}


\credit{Conceptualization of this study, Methodology, Software}

\affiliation[1]{organization={Department of Chemical Engineering, Norwegian University of Science and Thecnology},
    addressline={Gl\o{}shaugen}, 
    city={Trondheim},
    postcode={7034}, 
    country={Norway}}

\author[1]{Johannes J\"{a}schke}
\ead{johannes.jaschke@ntnu.no}



\author{I‪delfonso B. R. Nogueira}
\cormark[1]
\ead{idelfonso.b.d.r.nogueira@ntnu.no}


\cortext[cor1]{Corresponding author}



\begin{abstract}
The concept of creating a virtual copy of a complete Cyber-Physical System opens up numerous possibilities, including real-time assessments of the physical environment and continuous learning from the system to provide reliable and precise information. This process, known as the twinning process or the development of a digital twin (DT), has been widely adopted across various industries. However, challenges arise when considering the computational demands of implementing AI models, such as those employed in digital twins, in real-time information exchange scenarios. This work proposes a digital twin framework for optimal and autonomous decision-making applied to a gas-lift process in the oil and gas industry, focusing on enhancing the robustness and adaptability of the DT. The framework combines Bayesian inference, Monte Carlo simulations, transfer learning, online learning, and novel strategies to confer cognition to the DT, including model hyperdimensional reduction and cognitive tack. Consequently, creating a framework for efficient, reliable, and trustworthy DT identification was possible. The proposed approach addresses the current gap in the literature regarding integrating various learning techniques and uncertainty management in digital twin strategies. This digital twin framework aims to provide a reliable and efficient system capable of adapting to changing environments and incorporating prediction uncertainty, thus enhancing the overall decision-making process in complex, real-world scenarios. Additionally, this work lays the foundation for further developments in digital twins for process systems engineering, potentially fostering new advancements and applications across various industrial sectors.

\end{abstract}



\begin{keywords}
Digital Twin \sep Artificial Intelligence \sep Uncertainty \sep Transfer Learning \sep Online Learning 
\end{keywords}

\maketitle

\section{Introduction}

The concept of creating a virtual copy of a complete Cyber-Physical System (CPS) opens up a multitude of possibilities. This involves the capacity to virtualize the CPS in a comprehensive virtual environment, which allows for real-time assessments of the physical environment and vice versa \citep{TAO2019653,Eckhart2019,BIESINGER2019355,LO2021101297}. By constantly learning from the physical environment, it provides reliable and precise information about the actual scenario. This process, known as the twinning process or developing a digital twin (DT), represents virtual cloning \citep{9198575,HE2019221,Gao2022,8477101}. First proposed by Michael Grieves and John Vickers during a presentation at NASA in 2017 \citep{GrievesMJ2016}, NASA quickly became one of the pioneering enterprises to utilize the digital twin concept for space exploration missions. 

The concept of a digital twin revolves around creating a real-time link between a physical scenario and its virtual equivalent \citep{HAAG201864,LIU2021346}. A survey of relevant literature reveals that digital twins are widely regarded as a critical technology for the industries of the future \citep{Ganguli2020,Qi2021,Jones2020}. Since 2017, there has been a marked increase in the number of publications discussing the digital twin concept \citep{Jones2020}, with a significant portion focusing on manufacturing and product life cycle assessment domains. Digital twins are also gaining traction in the Process Systems Engineering field \citep{Melesse2021,systems7010007}. However, a primary challenge in these systems is the need to interactively solve complex numerical problems arising from partial differential mathematical models representing the systems.

Within the realm of digital twins, Artificial Intelligence (AI) plays a crucial role in facilitating the modeling, representation, and learning of complex behaviors and interactions among system components and data \citep{Kaur2020,9359733}. While phenomenological models can achieve similar outcomes, the extensive computational effort typically needed to solve these models numerically becomes impractical for real-time information exchange. Moreover, AI models offer the advantage of continuous learning from the system, thus providing the Cyber-Physical System (CPS) with adaptive capabilities. This approach, commonly called online learning, is an efficient tool and strategy to leverage the low computational effort needed for running an AI model online \citep{9359733,Gong2022}.

As a result, there is a burgeoning demand for research on the development and integration of AI and digital twins \citep{ZOHDI2020112907,Goodwin2022}. However, introducing online learning to the system increases the demand for computational power, which grows as the frequency of online learning activation increases \citep{Goodwin2022,SONG2022119995}. Therefore, resource management must be carefully implemented in such scenarios to preserve the benefits of using surrogate models. This challenge becomes more pronounced when considering the prediction uncertainty of AI models, which typically involves multiple evaluations of a probability distribution of the model's parameters \citep{Thelen2023,Gawlikowski2023,8371683}.

Overall, models developed using machine learning have become increasingly popular for performing inference and decision-making tasks in various fields. However, thoroughly evaluating their reliability and effectiveness is necessary to apply these Artificial Intelligence (AI) strategies in practice. The predictions generated by these models can be affected by noise and errors inherent to the inference and modeling methods used. Therefore, it is of utmost importance to consider AI models' uncertainty and possible limitations when making critical decisions based on their predictions. Therefore, it is highly desirable to represent uncertainty reliably in any AI-based system \citep{pawlowski2018implicit,math11010074}.

In response to these demands, there is a growing interest in developing models that are not only computationally efficient but also robust, adaptive, and endowed with a degree of cognition \citep{LIN2021101209}. Such models can self-adapt when discrepancies between their predictions and the current measured state are detected. The need for robustness in situations involving AI models is an increasingly prevalent topic in the literature. For example, \cite{Costa2022} proposed a robust learning methodology for uncertainty-aware scientific machine learning models, considering sources of uncertainty such as the absence of a theory, causal models, sensitivity to data corruption or imperfection, and computational effort. They applied this methodology to develop soft sensors for a polymerization process, demonstrating the identified soft sensors' resilience to uncertainties. \cite{Gneiting2007} introduced a methodology for calibrating the distribution of a known random variable, addressing issues related to non-deterministic variables and their impact on AI predictions. However, these studies primarily consider an offline environment with a virtually unlimited amount of computational resources available.

Furthermore, knowing a model's prediction uncertainty is crucial as it provides valuable information to assess its reliability and limitations. Additionally, prediction uncertainty enables informed decision-making based on the model's predictions. By evaluating the model's reliability based on its uncertainty, it is possible to identify situations in which it is more accurate or prone to errors. This information is critical for identifying areas where the model requires improvement or additional data collection may be necessary to enhance its accuracy \citep{10.1007/978-3-030-83723-5_6,Rahman2021}.
In particular, CPS systems enhanced by robust digital twins are increasingly in demand in the oil and gas industry, especially for offshore exploration where equipment is located hundreds of meters underwater and inaccessible \citep{9104682,KNEBEL2023109363,10.2118/195790-MS}. This necessitates reliable and precise systems operating under stringent economic, safety, and environmental constraints. Moreover, the ecological impact of accidents in exploration fields is critical, heightening the need for reliability and safety in these processes. For instance, gas-lift systems, an artificial lift technique used in the oil and gas industry to enhance the production of hydrocarbons from wells, present several sources of uncertainty that can affect decision-making and optimal operation.

Despite these advancements, there is a lack of reports in the literature addressing the robustness and uncertainty of digital twin strategies. Additionally, there is a dearth of studies examining the concise integration of techniques such as online learning, transfer learning, and robustness assessment when implementing a digital twin. In this context, the present work proposes a digital twin framework for optimal and autonomous decision-making applied to a gas-lift process that employs:

\begin{enumerate}
    \item \textit{Offline training}  to identify the ML models in an environment using computationally intensive data.
    \item \textit{Bayesian inference} to construct nonlinear model parameter probability distribution functions (PDFs).
    \item \textit{Monte Carlo simulations} to determine ML model prediction uncertainty.
    \item \textit{Transfer learning} to deploy the identified model and its corresponding uncertainty in an online environment distinct from its original training space.
    \item \textit{Reducing model space with statistical confidence} to alleviate the computational burden associated with online learning.
    \item \textit{Cognitive tack} to imbue the system with cognition, enabling awareness of its predictions and data received from the plant.
    \item \textit{Online learning }to update the model structure and correct any drift that the DT might identify during the operating campaign.
\end{enumerate}





\section{Methodology}
\subsection{Offline Digital Twin Identification}
In the methodology proposed in this work, offline identification serves as a foundational step in establishing the digital twin. It involves training machine learning (ML) models in a computationally intensive environment using historical and potentially large datasets. This process is performed offline due to the significant computational resources required. Hence, offline training allows the ML models to learn complex patterns and relationships within the data, ensuring high accuracy in their predictions. Once the ML models are trained, they are integrated into the digital twin framework, serving as the brain behind the digital twin's decision-making abilities. This section will provide the methodological details regarding the proposed offline identification step.

\subsubsection{Design of Experiments and Data Collection}
\label{Design_of_Experiments}
The initial step in developing artificial neural networks involves acquiring data pertinent to the process of interest. During this stage, it is crucial to gather a substantial amount of data to comprehensively represent the operational domain of the process while accounting for exceptions and boundary conditions within the problem domain. In this regard, diversifying data significantly enhances the quality of predictions made by machine learning models \citep{KLEIN1999569}.

In this study, a previously validated phenomenological gas lift model was employed to generate synthetic data. This data, produced through simulations or computational algorithms, offers a cost-effective alternative to real historical data when there is a scarcity of volume, quality, and variability. On the other hand, as this model has been previously validated in a pilot unit, it will be employed as a virtual plant in this study, utilizing a software-in-the-loop (SIL) approach to serve as an environment to emulate with reliability the online implementation of a digital twin.

Another essential aspect to consider during the data acquisition phase is the selection of input variables for the model. It is critical to carefully determine the combination of variables used to gather system output data. Inputs should be generated without cross-correlations, as these correlations can result in data discrepancies, obscure the process behavior, or even inflate the dimension of inputs and skew the neural network training. Furthermore, poorly distributed data can lead to overfitting of the identified models. In this context, Design of Experiments (DoE) \citep{Hicks1964-HICFCI} is a systematic approach to experiment design and data collection, with the goal of improving efficiency and accuracy. While DoE is commonly used in physical, chemical, and biological experiments, it has also been widely explored for acquiring data in the context of neural networks. This approach aims to optimize the information obtained from the input space while avoiding unintentional correlations and ensuring samples are uniformly distributed within the operating space of the system.

To achieve this, the latin hypercube sampling (LHS) \citep{doi:10.1080/00401706.1987.10488205, Owen1994} algorithm was utilized. LHS is a powerful technique for generating quasi-random samples that stratify data from a specified distribution and probability range. Its efficient stratification capabilities enable us to assess the full range of process behavior with fewer points than pure random sampling. Consequently, LHS allows for an efficient and representative sample of the multivariate parameter space.

Several works in the literature approach using LHS to generate synthetic data in different areas of knowledge, especially in the chemical and industrial processes, aiming at constructing neural networks.
In this way, using LHS to generate synthetic data can be a valuable tool to improve the accuracy and reliability of neural network models.

\begin{figure}[h!]
	\centering
		\includegraphics[scale=.08]{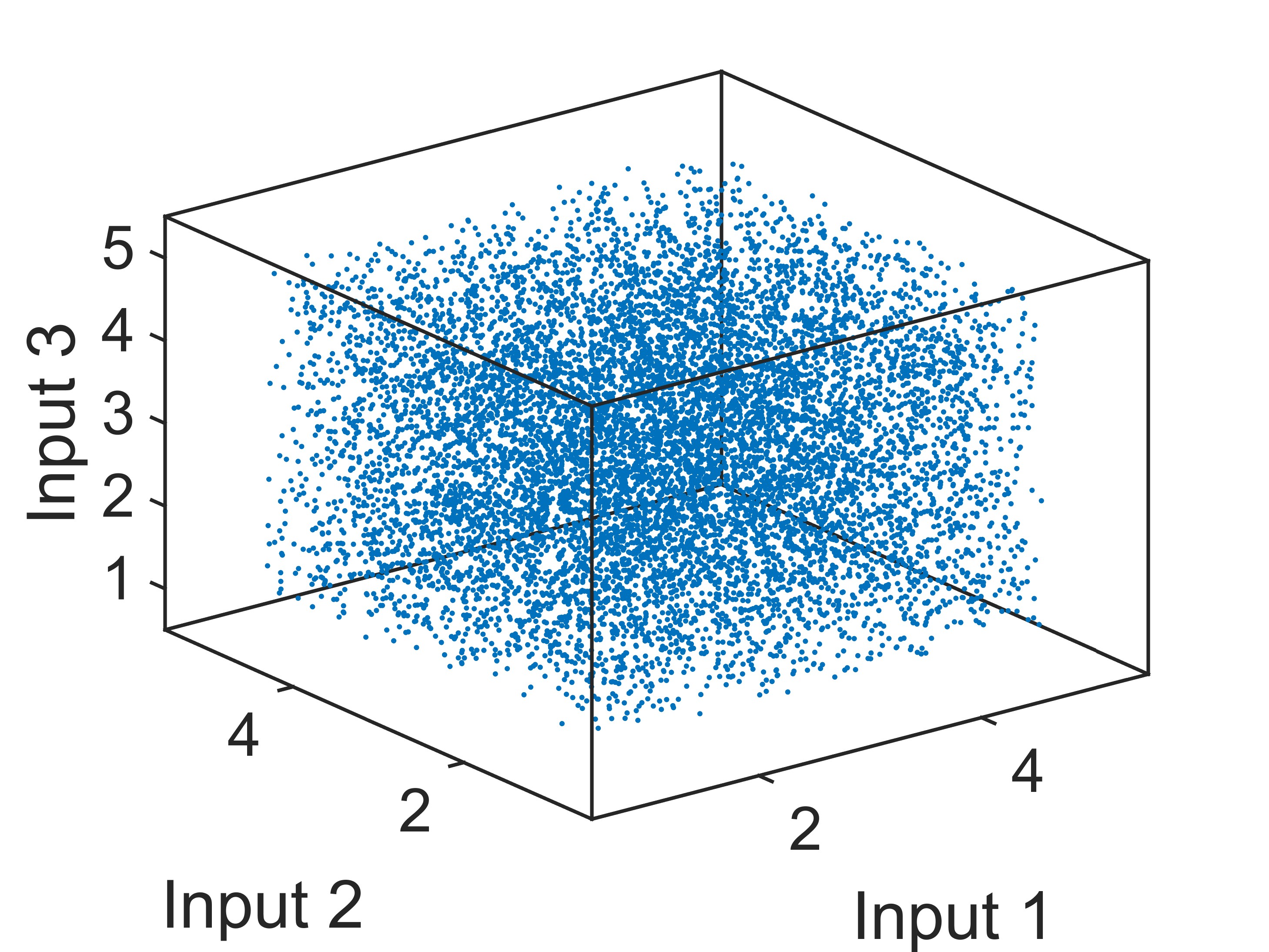}
    \caption{Latin hypercube sampling schematic representation}
    \label{lhs_FIG}
\end{figure}

The $lhsdesign$ function in MATLAB 2022b was used to implement LHS. The resulting input space was designed to extract the most information while minimizing unintended correlations. In Figure \ref{lhs_FIG}, shows a visual illustration of a 3D sample from the latin hypercube sampling method.

\subsubsection{Predictor structure identification}
\label{Predictor_structure_identification}

Choosing the appropriate data structure is a crucial step in any modeling approach. It involves identifying the predictor type and determining its corresponding embedding dimensions. One popular model for predicting nonlinear dynamic time series systems is the nonlinear autoregressive network with exogenous inputs (NARX), which \ first introduced cite{Leontaritis}. NARX has gained widespread adoption due to its effectiveness and versatility in modeling complex systems. In the realm of chemical and industrial systems, NARX networks demonstrate the efficiency and ability to identify long-term patterns 
\citep{Menezes2008,HangXie2009}.

NARX predictors are a form of predictor that incorporates feedback from the predicted output as input to the hidden layers during subsequent iterations. This feedback mechanism allows the network to model the temporal dynamics of input data and predict future output values while considering other influencing factors.

To identify a NARX predictor, a sequence of input data with corresponding exogenous inputs, delays, and their respective outputs are needed. The NARX prediction model can be enhanced by incorporating exogenous inputs, which helps it to capture dynamic behavior. This approach avoids overburdening the model's nonlinear function approximation with internal dynamics while still allowing it to track changes in the system. Additionally, the literature suggests that incorporating NARX structures with recurrent models can improve model predictability and lead to a more streamlined nonlinear model. This approach has been shown to be effective in enhancing the performance of AI models in various process systems applications \citep{Rebello2022,Nogueira2018}. By reducing the number of layers and weights needed, this approach reduces computational costs, making NARX predictors a practical choice for online applications.

Given these benefits, we used the NARX predictor in our work. Its ability to incorporate exogenous inputs and accurately capture system dynamics while keeping a small nonlinear model structure is an ideal choice for constructing a digital twin.

The mathematical expression for a NARX network can be represented as Equation \ref{eq_NARX}:
\begin{equation}
    \hat{y}(t)=f(y(t-1),y(t-2),\ \ldots,\ y(t-N_b\ ),u(t-1),u(t-2),\ \ldots,\ u(t-N_a)+e(t),
    \label{eq_NARX}
\end{equation}

\noindent where $y(t)$ represents the desired output variable, $\hat{y}(t)$ is the predicted output, $u(t)$ is the model's input variable, $N_a$ and $N_b$ are the predictor embedding dimensions, defined as the input, and output variable time delays, and $e(t)$ is the additive error.

The performance of a predictor is independent of the choice of the type of nonlinear function approximator to be used. Therefore, defining the predictor's embedding dimensions $(N_a)$ and $(N_b)$ is an essential step. Despite the crucial role of predictor parameters in accurately identifying dynamic systems, their definition and estimation are often overlooked in AI modeling literature. This oversight can lead to inaccuracies and errors in the modeling process. Therefore, it is important to emphasize the significance of defining and estimating predictor parameters in AI modeling to ensure the reliability and validity of the results. In this study, the Lipschitz coefficient $(q_j^{(n)})$ proposed by  \cite{He1993} was used to characterize the embedded nonlinear relationship between inputs and outputs of a complex dynamic system and identify the predictor embedding dimensions. The Lipschitz coefficient is calculated by the ratio of the difference between function output ($y$) values and the distances between the respective inputs ($x$), according to Equation \ref{eq_lipchitz}.

\begin{equation}
    q_j^{(m)}=\frac{\left|\delta y\right|}{\sqrt{(\delta{x_1)}^2+...+(\delta{x_{m})}^2}}=\frac{\left|f_1\delta x_1+...+f_m\delta x_m\right|}{\sqrt{(\delta{x_1)}^2+...+(\delta{x_{m})}^2}},
    \label{eq_lipchitz}
\end{equation}

where $m$ represents the number of input variables in the input-output formulation. 

The Lipschitz Index is used to identify the ideal number of delays. This index is calculated according to Equation \ref{index_lipschitz}:

\begin{equation}
    q^{(n)}=\left(\prod_{k=1}^{p}\sqrt{n}q_j(k)^{(m)}\right)^{\left(\frac{1}{p}\right)},
    \label{index_lipschitz}
\end{equation}

\noindent where $n$ is the number of delays considered in the variables, $p$ is a parameter usually between 0.01 N and 0.02 N and ${q(k)}^{(n)}$ is the k-th most significant Lipschitz coefficient from all $q_j^{(m)}$ calculated in Equation \ref{eq_lipchitz}

The method consists of testing different values of the delay number, represented by $n$, and calculating the value of the Lipschitz index for each tested delay value. The goal is to determine if there is a significant difference between the Lipschitz index values calculated for different values of $n$. Based on these calculations, it is possible to identify the first index that indicates a region in which variations of $n$ do not significantly affect the calculated value of the Lipschitz index. This index corresponds to the ideal number of delays desired for the inputs.

It is essential to highlight that using Lipschitz metrics in constructing and analyzing mathematical models, whether for machine learning or other types, is important for providing information that ensures stability and convergence of training optimization algorithms and model adjustments.

This work adopted the Multiple-Input Single-Output (MISO) \citep{8756639} strategy due to its ease of identification and real-time training for neural networks.

\subsubsection{Nonlinear model Hyperparameters Identification}
\label{Hyperparameters}

After collecting and defining the predictors, the next step is to define the hyperparameters that shape the nonlinear model's structure. These hyperparameters can be broadly categorized into two groups: model parameters and algorithm parameters.
Model parameters, which are established before training begins, determine the network's architecture and include the number of layers, the number of neurons per layer, the layer type, the initial learning rate, the batch size, the number of epochs, and other essential features. These parameters are crucial in identifying a good neural network, as they directly impact the model's function, structure, and performance.
The set of hyperparameters comprises discrete and continuous variables, which makes the appropriate choice challenging, considering the number and type of variable.

In contrast, algorithm hyperparameters are the internal parameters that are updated during the learning process. These parameters are adjusted during training, such as regularization parameters, learning rate schedules, momentum, and optimization algorithms. Proper tuning of these hyperparameters can help the network generalize better and avoid overfitting.

In the field of machine learning, it is commonplace to use trial and error methods, such as random search and grid search, for hyperparameter tuning \citep{10.5555/2188385.2188395,JMLR:v18:16-558}. However, these methods are computationally expensive and inefficient, making them less than ideal.

A promising alternative is the HYPERBAND method, which has gained significant attention in hyperparameter optimization due to its superior efficiency and precision. HYPERBAND is an optimization algorithm that utilizes random sampling and early stopping of model training to minimize the number of evaluated hyperparameter combinations. This method discards low-performing models while allowing high-performing models to continue in the optimization process. As the search continues, resources are gradually allocated to the most promising models until the best hyperparameters in the search space are identified.

Overall, HYPERBAND's resource allocation strategy optimizes hyperparameters efficiently, reducing the computational cost and time associated with traditional methods like random search and grid search. Its popularity among machine learning practitioners is on the rise, and it is expected to play an essential role in future hyperparameter optimization studies.

Before using the HYPERBAND optimization algorithm, a preliminary step is to define the hyperparameters of interest and their corresponding search spaces. These hyperparameters may include the learning rate, the number of neural network layers, the batch size, and other parameters that can impact the model's performance. Each hyperparameter is defined within a specific search space, which usually represents a range of possible values for that parameter. By defining these search spaces, the HYPERBAND algorithm can explore the different combinations of hyperparameters and find the optimal set that produces the best model performance. 

Choosing an appropriate search space and parameter set is important to achieve more precise and computationally efficient results. It is necessary to balance the search for space exploration and the intensification of training in specific areas. Otherwise, the algorithm may save time and effort on parameters that will not significantly impact the model's performance. Furthermore, it would increase the probability of overfitting. For this reason, in this work, the hyperparameter's Initial learning rate, number of dense layers, activation function in each layer, and number of neurons in each layer were selected to find the optimal set of parameters for the model. 

In the present study, the hyperparameter search space is represented in Table \ref{TAB:Hyperparameters}.

\begin{table}[width=.6\linewidth,cols=2,pos=h]
\renewcommand{\arraystretch}{1.3} 
\caption{Hyperparameter search space for HYPERBAND}
\begin{tabular*}{\tblwidth}{@{} LLL@{} }
\toprule
Hyperparameters & Search space \\
\midrule
Initial learning rate & $1\times{10}^{-3}$,$1\times{10}^{-2}$,$1\times{10}^{-1}$		\\
Number of dense layers	& \{1,2,3,4\} 	\\
Activation function in each layer & \{relu, tanh\}	 \\
Number of neurons in each layer  & from 20 to 100 with 10 step		\\

 \bottomrule
\end{tabular*}
\label{TAB:Hyperparameters}
\end{table}

To fine-tune a neural network model using the HYPERBAND algorithm, it's crucial to have access to the training, validation, and test sets. In the present methodology, these sets were obtained during the data acquisition phase and organized based on the chosen predictors using the Lipschitz Index.

During the optimization process, the HYPERBAND algorithm iteratively tunes the model's hyperparameters to improve its performance with respect to the defined objective function. This objective function is typically evaluated using the training and validation data with the hyperparameters chosen within the specified search space. The objective function in the present case includes a loss function that measures the model's performance on both the training and validation sets.

\begin{figure}[h!]
	\centering
		\includegraphics[scale=.6]{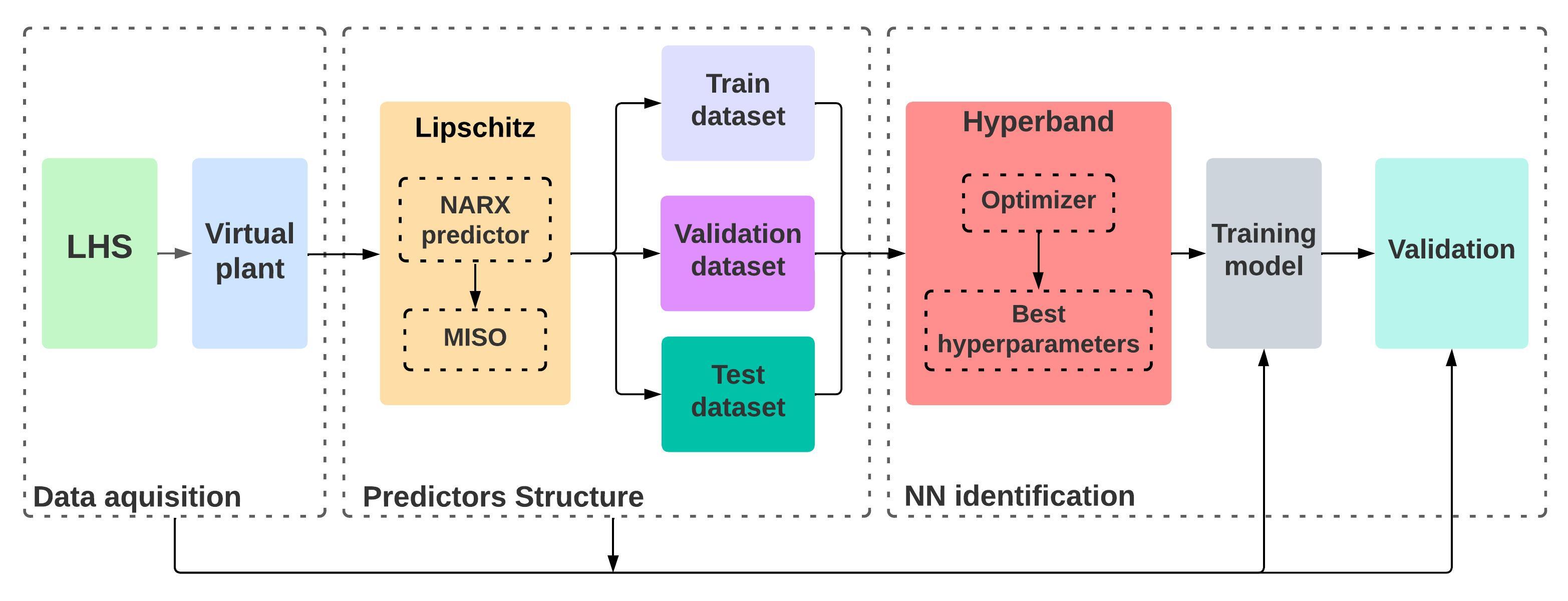}
    \caption{Flowchart of the nonlinear model identification strategy presented in this work}
    \label{metodologia_nn}
\end{figure}

After completing the optimization process, the model's performance is assessed using the test data to determine its ability to generalize to new datasets. The model with the optimal hyperparameters can then be chosen as the final model for the uncertainty assessment. Figure \ref{metodologia_nn} presents a schematic representation of the methodology presented in this section.

\subsubsection{Markov Chain Monte Carlo Uncertainty Assessment}
\label{incerteza}

During neural network training, an optimization process is performed to adjust various parameters known as weights and biases. This process involves the repetition of several epochs, in which the weights are adjusted to achieve the desired performance of the neural network. Optimization of weights is often performed using techniques such as stochastic gradient descent, which adjusts weights based on prediction error. Once the neural network has been trained, the final weights are used to make predictions on new input data. As these weights are stochastic variables, they will have an associated probability distribution, making it possible, even though usually neglected, to identify the uncertainty associated with the parameters and the propagation for model prediction.

Overall, the literature presents several methods to evaluate uncertainty in model predictions. Among these methods, the Bayesian method combines information from an a priori probability distribution with sample information produced in a posterior probability distribution of one or more parameters in a parametric space. This Bayesian approach offers a more comprehensive and complete view regarding meditation, allowing the inclusion of previous information about the parameters. In contrast, a frequentist approach, which might include simplifications such as the Least Squares and Maximum Likelihood Methods, does not provide a probability distribution for the parameters but instead assigns a fixed value to them.
Therefore, in this work, we chose Bayesian inference to be used in the proposed methodology for identifying robust digital twins.

In the Bayesian approach \citep{Finkelstein1970,LAMPINEN2001257}, the true value of the parameters $\boldsymbol{\theta}$ is unknown. Therefore, it is possible to quantify the uncertainties associated with the values of $\boldsymbol{\theta}$ in terms of probability distributions $(P(\boldsymbol{\theta}))$. This approach is advantageous because it allows for the incorporation of prior information about the parameters before data acquisition by assigning a probability distribution.
However, when no prior information is available or a more conservative scenario is desired, a non-informative prior can be used. Thus, ensuring that the posterior distribution is not influenced by unreliable or subjective information.

Once the prior is defined, the likelihood function is determined to obtain the posterior density distributions of $\boldsymbol{\theta}$ so that any information regarding the parameters $\boldsymbol{\theta}$ can be obtained from the posterior probability density function (PDF). The process of Bayesian inference involves using reference data to update the prior probability distribution to obtain the posterior probability distribution. This is achieved by applying Bayes' theorem \citep{Swinburne2004-SWIBT-2, Koch1990} represented in Equation \ref{Bayes}.

\begin{equation}         
\mathbf{g}_{\boldsymbol{\theta}}\boldsymbol{(\left.\eta\right|D,I)}\propto L\boldsymbol{(\left.\eta\right|D)}\boldsymbol{g_\theta(\left.\eta\right|I)},
\label{Bayes}
\end{equation}

\noindent where $\eta$ represents sampled values of $\boldsymbol{\theta}$, $L$ is the likelihood function, $\boldsymbol{g_\theta(\left.\eta\right|I)}$ are the
prior distributions of $\boldsymbol{\theta}$ that are a new observation of $\boldsymbol{\theta}$, and $\boldsymbol{g_\theta(\left.\eta\right|D,I)}$ represents the posterior probability distribution.

In this work, the likelihood function used was the Mean Squared Error (MSE) as approved Equation \ref{likelihood}:

\begin{equation}   
L\boldsymbol{(\eta\mid D)} = \frac{1}{n}\sum_{i=1}^{n}(y_{i}-\widehat{y_i})^{T}(y_{i}-\widehat{y_i}),
\label{likelihood}
\end{equation}

\noindent where $y_{i}$ is the $ith$ observed value, $\widehat{y_i}$ is the corresponding predicted value, and $n$ is the number of observations.
The posterior PDF of each parameter $\theta_i$ of the vector $\boldsymbol{\theta}$, $\boldsymbol{g_\theta(\left.\eta\right|D,I)}$ are obtained from the marginal posterior density function $\boldsymbol{g_\theta(\left.\eta\right|I)}$ and this is defined in the Equation \ref{posterior}:

\begin{equation}   
\boldsymbol{g_\theta(\left.\eta\right|D,I)}\propto\ \int_{np-1}{L\boldsymbol{(\left.\eta\right|D)g_\theta(\left.\eta\right|I)}d\boldsymbol{\theta}_{np-j}}.
\label{posterior}
\end{equation}

Identifying the posterior PDF of each parameter involves solving the inference problem composed of Equations \ref{likelihood} and \ref{posterior}. Therefore, the posterior probability distribution represents the updated belief about the unknown parameter after incorporating the experimental data. It combines the prior probability distribution and the likelihood function, following the parameter values that are both supported by the prior beliefs and consistent with the observed data.

Especially in complex nonlinear models, the solution cannot be obtained analytically, requiring numerical estimation. For this task, the Monte Carlo method via Markov Chains (MCMC) \citep{Brooks1998} is a valuable technique in the Bayesian inference context, as it enables the problem's solution through sampling from the posterior distributions of the parameters of interest. 

\begin{figure}[h]
	\centering
		\includegraphics[scale=.6]{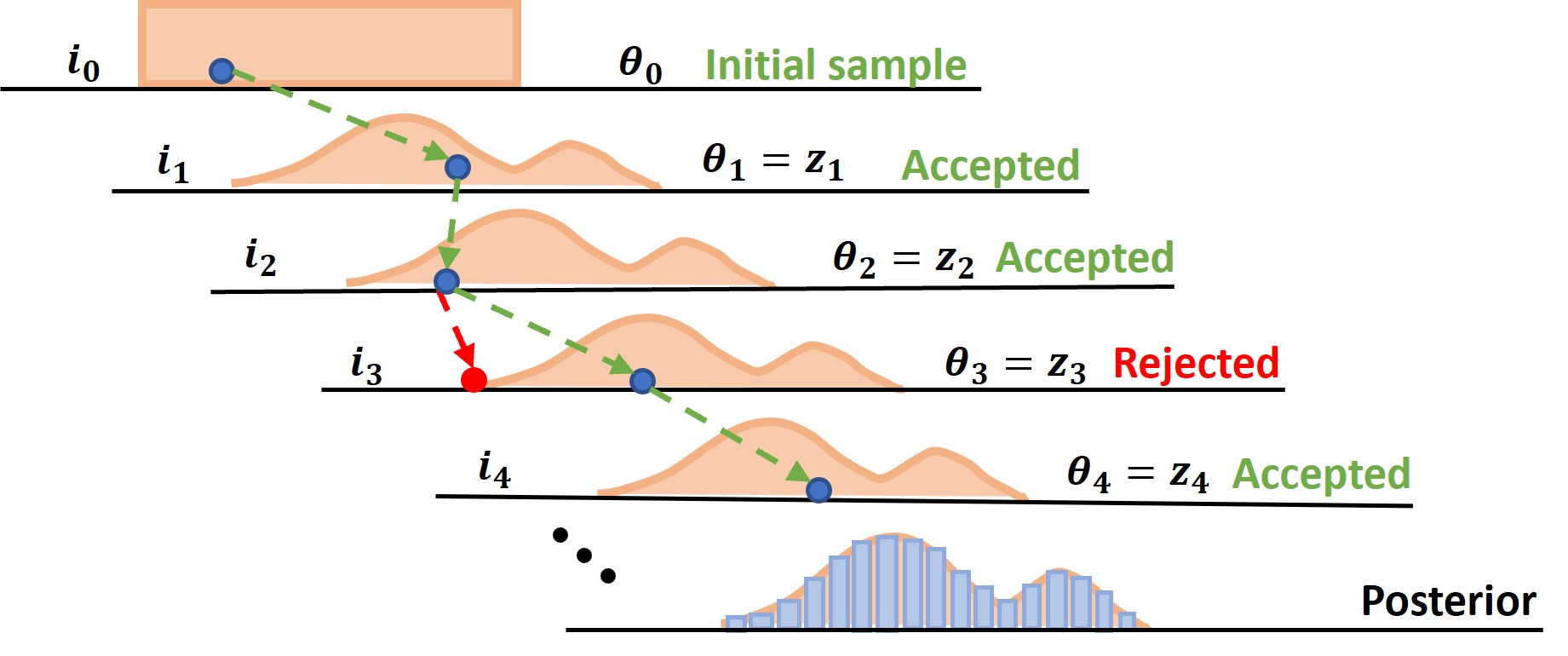}
    \caption{Markov Chains Monte Carlo method schematic representation}
    \label{MCMC.}
\end{figure}

The MCMC is used in this context as an iterative method that generates random samples from a proposed distribution to estimate the posterior distribution. At each iteration, the proposed distribution generates a new sample, which is accepted or rejected based on a reception probability determined by the relationship between the new sample's posterior density and the current sample's posterior density, as shown schematically in Figure \ref{MCMC.}.

By repeating this process, the MCMC generates a sequence of random sequences of the posterior distribution of the parameters of interest, allowing the belief of their marginal distributions and the identification of possible correlations between the parameters.

In this study, the most probable value $\boldsymbol(\hat{\theta})$ for each parameter of the neural networks was calculated according to the following Equation \ref{valor_provavel},

\begin{equation}   
 \hat{\theta}=\int_{-\infty }^{\infty }\eta \boldsymbol{g_{\theta}(\left.\eta\right|D,I)}d\boldsymbol{\theta},
 \label{valor_provavel}
\end{equation}

\noindent and, the covariance matrix $(U_{\theta \theta })$ of the parameters is defined by the following Equation \ref{matrix_cov}: 

\begin{equation}   
U_{\boldsymbol{\theta \theta}} = \int_{-\infty }^{\infty }(\eta -\hat{\theta })^{T}(\eta -\hat{\theta })\boldsymbol{g_\theta(\left.\eta\right|D,I)}d\boldsymbol{\theta}.
\label{matrix_cov}
\end{equation}

Once the PDF of the parameters of neural networks is built, it is possible to propagate the uncertainty of the parameters to the prediction using techniques such as Monte Carlo (MC) sampling \citep{SHAPIRO2003353}. The methodology consists of selecting random samples of parameters from their probability distributions and performing several predictions with these different groups of parameters. From these predictions, it is possible to calculate the probability distribution of the response and, therefore, the uncertainty of the final prediction, incorporating the uncertainty of the parameters.

\begin{figure}[h!]
	\centering
		\includegraphics[scale=.57]{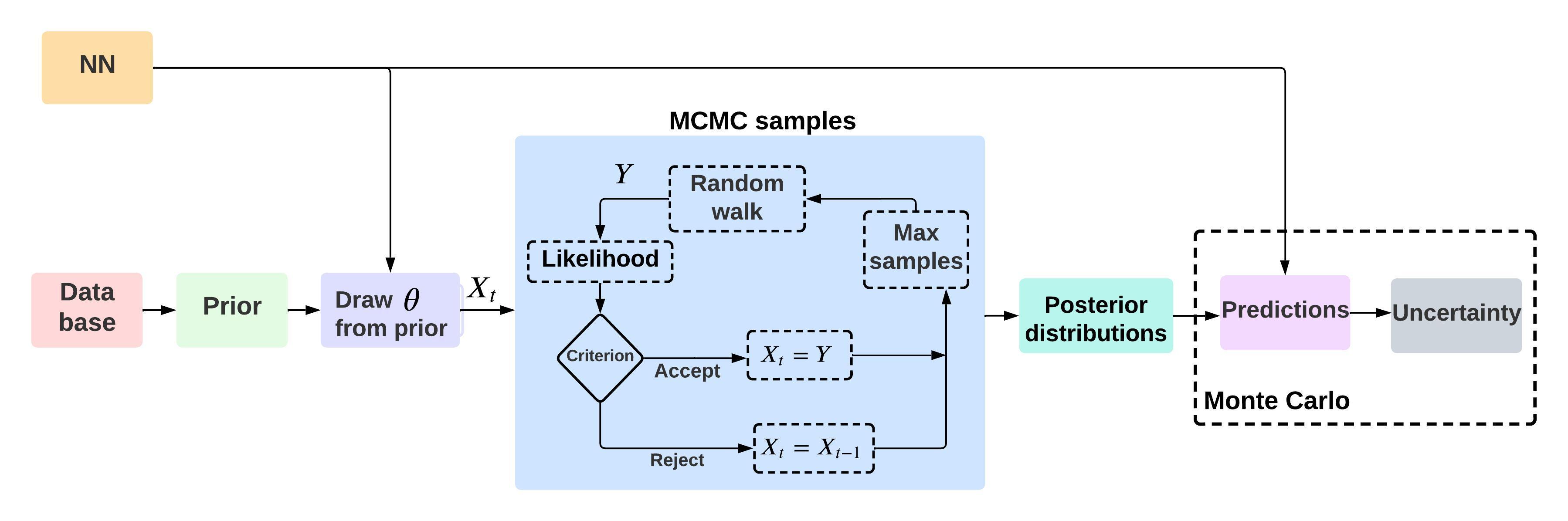}
    \caption{Flowchart of the uncertainty strategy presented in this work}
    \label{Metodologia_offline}
\end{figure}

In Figure \ref{Metodologia_offline}, a schematic diagram of the methodology described in this section is presented.

\subsection{Online Digital Twin Implementation}

In the offline training phase, one can use computationally expensive data without concerns regarding the demand for online feedback. However, when transitioning to an online environment, computational resources are limited by the frequency at which predictions are required. The transfer learning strategy addresses this issue by utilizing computationally expensive information in the offline context and bringing it to an online context by transferring previously acquired knowledge. Hence, the first step to implementing an online digital twin is to transfer the knowledge and its corresponding uncertainty acquired in an offline environment to an online environment. 

In this context, it is crucial to capitalize on the knowledge acquired during the offline phase of the digital twin and use it as a foundation for deploying the digital twin (DT) in an online setting. This technique, known as transfer learning in machine learning, is employed in this work due to its advantages. These benefits include reducing the amount of training data required for online learning purposes, decreasing the computational effort associated with the structural identification of neural networks (by using transfer learning, it can be assumed that the model structure identified during the offline phase is optimal), and significantly enhancing the performance of the new model. By utilizing transfer learning, the online digital twin effectively leverages offline knowledge while adapting to real-time data and varying operational scenarios, thus ensuring a seamless transition and efficient performance in the online environment.

A subsequent step within the online environment is the Reducing model node. This is meant to reduce the hyperspace of probable models and project it into a low-dimensional space that comports the computational resources of the online environment. This theorem is presented in Equation \ref{red_dim}.

\begin{equation}   
R^{n}\rightarrow R^{n-q}, 
\label{red_dim}
\end{equation} 

\noindent where $n-q$ is given by the sensitivity analysis.

In this scenario, $R$ represents the n-dimensional space of the models, while $(n-q)$ corresponds to the low-dimensional space. Determining the dimension reduction factor $q$ is critical, as it impacts the online performance of prediction uncertainty. In this case, the factor was identified through an offline sensitivity analysis. The value of $q$ was successively reduced until the prediction uncertainty fully degenerated. The degeneration inflection point can serve as the minimum position dimension. This study introduced a safety factor of 25\% to the methodology, ensuring that the digital twin (DT) operates well away from the degeneration point. After defining the dimension reduction factor $q$, the model parameters' original probability density function (PDF) is randomly sampled to populate the new reduced space. This PDF dimension reduction and sorting strategy aims to minimize the computational effort of running a large distribution of potential models online. Subsequently, the uncertainty in the parameters can be propagated to the prediction using MC methods. This process involves selecting random samples of parameters from their respective probability distributions and conducting multiple predictions with these diverse parameter sets. From these predictions, the probability distribution of the response can be calculated, enabling the determination of the uncertainty in the final online prediction while accounting for the uncertainty of the parameters.

\begin{figure}[h!]
	\centering
		\includegraphics[scale=.57]{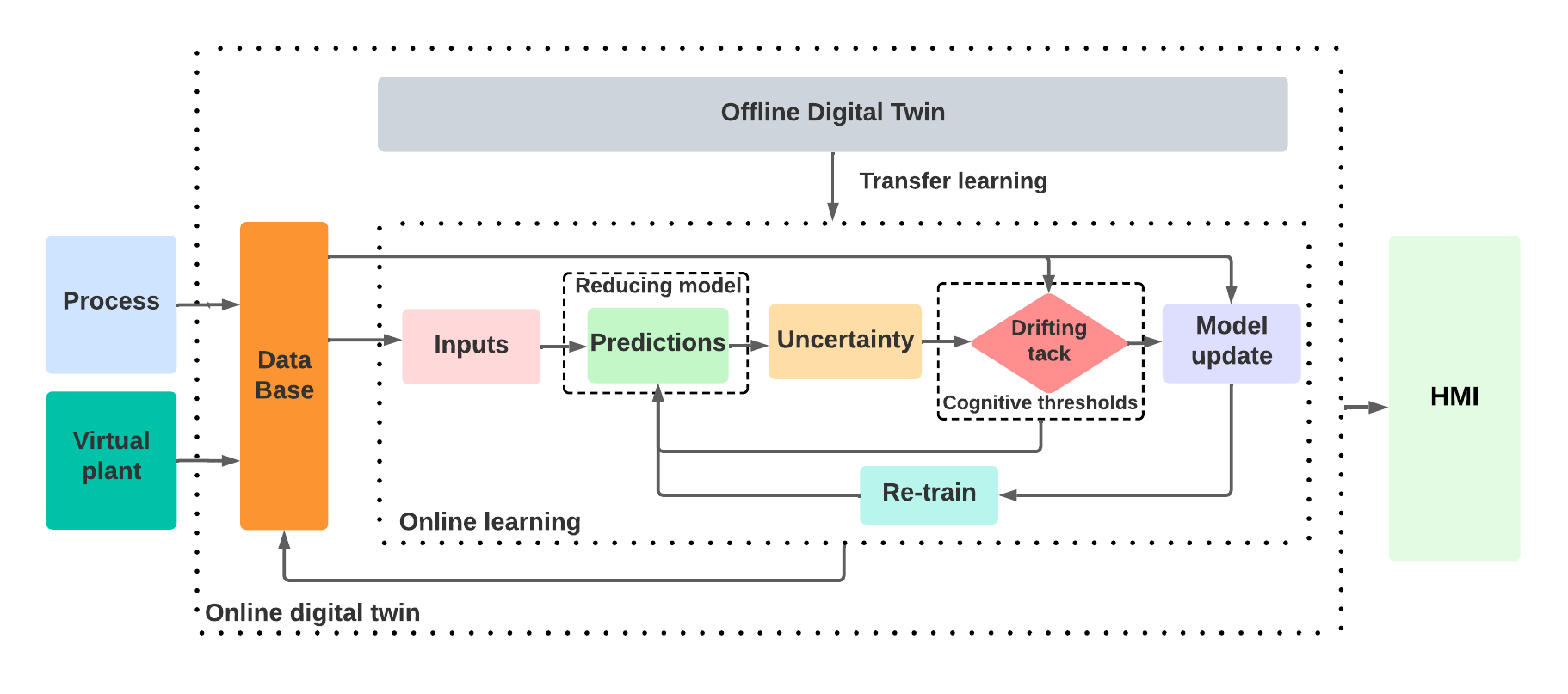}
    \caption{Flowchart of the online digital twin strategy presented in this work}
    \label{metodologia_online}
\end{figure}

The next step is the self-awareness component proposed in this work. The necessity for a digital twin to be self-aware of the quality of its predictions in relation to the system's current state is vital for several reasons. As the digital twin might play a critical role in monitoring, predicting, and optimizing the system, their effectiveness depends on the accuracy and reliability of their predictions. Hence, this work proposes a self-aware digital twin that can better adapt to system changes as it continually evaluates its predictive performance. This identifies potential discrepancies between the model and the real system, enabling real-time adjustments to improve prediction accuracy \citep{doi:10.1080/00207543.2021.2014591,9474166}. This is done within the cognitive track block and the cognitive node, Figure \ref{metodologia_online}. The self-aware digital twin fosters a continuous learning environment where the model learns from its successes and failures, refining its predictions over time. Hence, the cognitive node will be the instance responsible for controlling the online learning activation. When the cognitive threshold (CT) is achieved, the node triggers online learning to update the digital twin. This results in a more robust and reliable model that can adapt to various operational scenarios. Equations \ref{cognitive_1} and \ref{cognitive_2} present the functions behind the cognitive node, which were developed inspired by the activation mechanism of a neuron and considering the prediction uncertainty of the DT.

\begin{equation}   
Z = \sum_{n=a}^{b} [H(y_{mesured} - Inf(y_{Coverage Region}) + H(Sup(y_{Coverage Region})-y_{measured})],
\label{cognitive_1}
\end{equation} 

\noindent where $H$ is a Heaviside function. The infimum ($\inf$) and supremum ($\sup$) operators will compute the smallest lower bound and least upper bound of the DT's coverage region, respectively. When the measured value falls outside the bounds of the coverage area, the Heaviside function activates and returns a value of one. However, the Heaviside function remains inactive if the predicted value is within the coverage area. Essentially, this means that the measurement is located within the coverage regions of the DT. $a$ and $b$ are functions of a moving horizon (MH) factor, which can be computed by:

\begin{equation}
    \begin{cases}
     & a = 0 + k   \\ 
    & b = MH + k.
    \end{cases}
    \label{cognitive_2}
\end{equation}

The MH factor is a component that aids the cognitive tracker in maintaining an accurate understanding of the digital twin's current state. This dynamic approach ensures that the digital twin remains responsive and adaptive to real-time changes in the system it represents, enhancing its overall effectiveness. The moving horizon factor continually updates the analyzed data window, ensuring that the most recent information is always considered. This approach allows the cognitive tracker to focus on the most relevant data and discard older, less pertinent information. As a result, the digital twin can effectively respond to changes in the system and maintain an accurate representation of its current state. Overall, this adds a memory component to the cognitive node.

Furthermore, in the real-time online environment, disturbances and unforeseen scenarios different from those encountered during offline training may occur. Therefore, the digital twin must receive real-time performance data from the system and evaluate the necessity of incorporating it into its learning to make reliable predictions. Since industrial processes are highly dynamic and subject to various sources of disturbances and unpredictable scenarios, it is crucial that the digital twin can identify behavior changes, adapt quickly to new system scenarios, and incorporate new data in real time to update the tool. This makes it possible to obtain increasingly accurate and reliable predictions. To achieve this goal, in this work, the online learning tool was integrated into the digital twin to enable the system to incorporate cognition to identify scenario changes and constantly check the predictions and data that the process receives. The implementation of this online learning tool is building a new database as the measurements are collected and processed so that the digital twin can retrain the neural networks when necessary. Therefore, the online learning methodology updates the model and corrects possible deviations identified by the digital twin during the operational campaign. 

Figure \ref{metodologia_online} presents a schematic representation showcasing the integration of essential concepts that form the proposed digital Twin foundation. This work specifically combines transfer learning, uncertainty management, hyperdimensional reduction techniques for parameter selection and PDF construction, system awareness of changes in plant scenarios (through cognitive nodes and cognitive thresholds), collaborative online learning concepts, and finalizing the strategy with the results presented in a man-machine interface (HMI).

It is important to highlight that this approach accommodates synthetic and real data. Synthetic data generated by phenomenological models serve a dual purpose: they not only increase the volume of data required for effective neural network training but also facilitate the generation of risk scenarios or operational abnormalities that might not be frequently encountered during daily operations \citep{7966298}. Meanwhile, process data can be seamlessly incorporated into the digital twin system following proper data curation, which is simulated in this work. As a result, Figure \ref{metodologia_online} illustrates both potential data sources for constructing the dataset, emphasizing the versatility and adaptability of the proposed digital Twin Online approach.

\section{Case Study: Gas Lift System}
\subsubsection{Gas Lift System}
A gas lift unit consists of a method for the artificial lift of oil and gas used to boost the production of hydrocarbons from wells with insufficient natural pressure. This technique's basic concept consists of injecting compressed gas, typically natural gas, into the wellbore, which reduces the density of the fluid column and creates a gas-oil mixture that is easier to lift to the surface. In this study, the use of a gas lift pilot unit was considered. In this study, the use of a gas lift pilot unit was considered. This unit is a small-scale experimental platform that simulates different underwater oil well network processing scenarios.

\begin{figure}[h!]
	\centering
		\includegraphics[scale=.65]{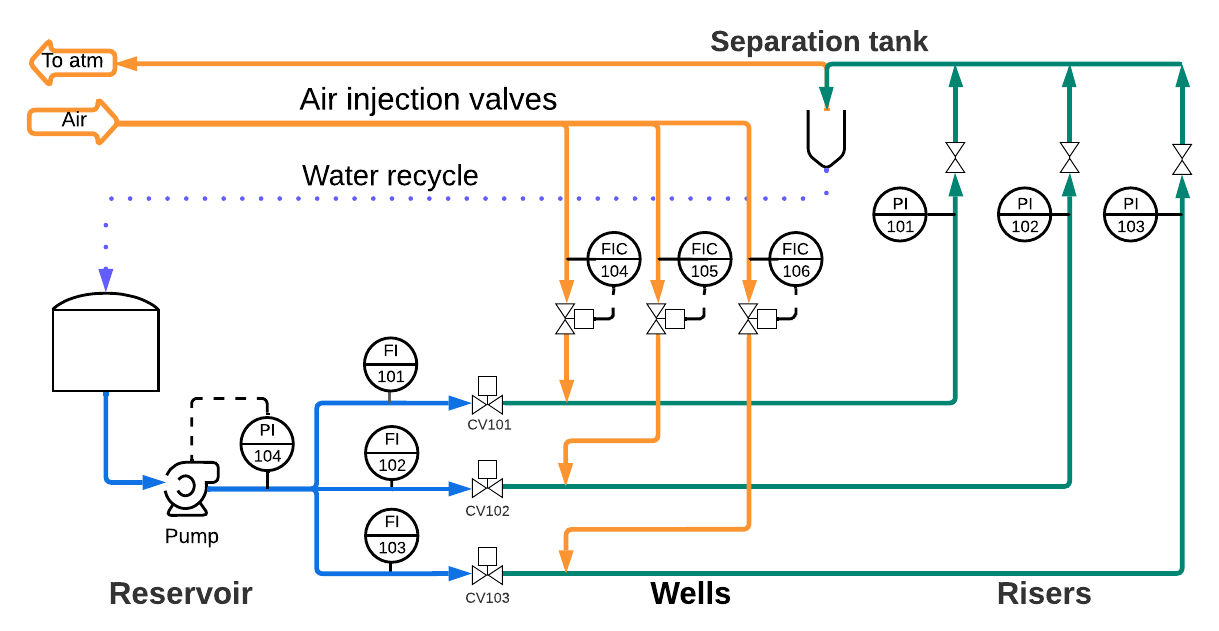}
    \caption{Experimental setup of gas lift experiment adapted from \cite{Matias2022}. The measured system variables are the well top pressures (PI101, PI102, and PI103), the pump outlet pressure (PI104), the liquid flowrates (FI101, FI102, and FI103), and the gas flowrates (FI104, FI105, and FI106). The reservoir valve openings (CV101, CV102, and CV103) are the system disturbances.}
    \label{gas_lift}
\end{figure}

Figure \ref{gas_lift} shows the didactic division of the industrial prototype into three sections: reservoir, wells, and risers. In this system, the working fluids are water and air, replacing oil and air. The reservoir consists of a 200 L steel tank, centrifugal pump, and control valves (CV101, CV102, and CV103). In this work, the valve openings were used to simulate different behaviors resulting from the reservoir, which only produces liquid. As shown in Figure \ref{gas_lift}, flow meters (FI101, FI102, and FI103) are located before the reservoir valves.

The simulation of the wells in the experimental prototype is performed through three flexible hoses with a diameter of 2 cm and a length of 1.5 m. The gas lift system air is injected 10 cm after the CV101, CV102, and CV103 valves, within the range of 1 to 5 $sLmin^{-1}$. A control system can be activated (FIC104, FIC105, and FIC106) to regulate the injected gas flow in the system.

Finally, the risers are represented by three vertical tubes with an internal diameter of 2 cm and a height of 2.2 m, which are orthogonal to the well-representative pipes. The pressure at the top is measured by the gauges PI101, PI102, and PI103, and three manual valves are located in sequence, which is kept open during the experiments. At the end of the process, the liquid is recirculated in a closed system and returned to the reservoir, while the air is expelled into the atmosphere.

In the current study, we employed a first principle model of the gas lift process. This model functions as a virtual plant, providing a platform for developing and testing the proposed methodology. As the model has been previously proposed and validated in the literature, it offers a reliable source of information for our investigation \citep{Matias2022}. 

By employing this well-established and validated model, we are able to simulate various gas lift scenarios and analyze the performance of the proposed methodology under different operating conditions. Hence, facilitating the development process. 


The phenomenological model describing this system consists of a set of algebraic and differential equations based on \cite{Krishnamoorthy2018} model.
The model accounts for both hydrostatic pressure and pressure loss due to friction. When making calculations, the pressure difference along the riser is deemed insignificant, which means that only two pressures are considered: one at the bottom and another at the top of the riser \cite{Matias2022}.

The following equations represent the mass balances for liquid and gas in the system, known as differential Equations \ref{diferencial_1} and \ref{diferencial_2}:

\begin{equation}
    \dot{m}_{g} = w_{g} - w_{g,out}, 
    \label{diferencial_1}
\end{equation}

\begin{equation}
\dot{m}_{l} = w_{l} - w_{l,out},
\label{diferencial_2}
\end{equation}

\noindent where $\dot{m}_l$ and $\dot{m}_g$ represent the respective mass holdups of liquid and gas inside the wells and riser. $w_g$ represents the mass flowrate of gas injected into the system, while $w_l$ represents the flowrate of liquid coming from the reservoir. Additionally, $w_{g,out}$ and $w_{l,out}$ respectively denote the outlet production rate of gas and liquid from the system.

The algebraic equations are used to describe certain relationships within the system. The following equation can express the outflow from the reservoir:

\begin{equation}
    w_{l} = v_{o}\theta_{res}\sqrt{\rho_l(P_{pump}-P_{bi})},
\end{equation}

\noindent where $\theta_{res}$ is the reservoir valve flow coefficient, $\rho_l$ represents the density of the liquid in the system, and $v_o$ is the valve opening. The pump outlet pressure, $P_{pump}$, is measured, and the pressure before the injection point, $P_{bi}$, is calculated using hydrostatic pressure and accounting for the pressure drop due to friction. To simplify this calculation, we utilize the Darcy-Weisbach equation for laminar flow in cylindrical pipes. Therefore, the expression for $P_{bi}$ becomes:

\begin{equation}
    P_{bi} = P_{rh} + \rho_{mix}g\Delta h + \frac{128\mu_{mix}(w_g+w_l)L}{\pi\rho_{mix}D^4},
\end{equation}

\noindent where, $P_{rh}$ represents the pressure measured at the riser head, $\Delta h$, which denotes the height from the bottom of the well to the top of the riser, $L$, the length of the pipes (i.e., the combined length of the well and riser), $D$, the diameter of the pipes, $g$, the gravitational acceleration; and $\mu_{mix}$, the viscosity of the mixture of liquid and gas. In the experimental setup, the mixture viscosity is approximated by the liquid viscosity. The mixture $(liquid + gas)$ density $\rho_{min}$ is calculated by Equation \ref{rho}:

\begin{equation}
    \rho_{mix} = \frac{m_{total}}{V_{total}} = \frac{m_g + m_l}{V_{total}}.
    \label{rho}
\end{equation}

Additionally, there is an equation that states the sum of the volumetric holdups of gas $(V_g)$ and liquid $(V_l)$ is equal to the total volume of the system.

\begin{equation}
    {V_{total}} = V_g + V_l = \frac{m_l}{\rho_l} + \frac{m_g}{\rho_g}.
\end{equation}

The liquid density is considered to be constant, denoted by the symbol $\rho_l$. The gas density ($\rho_g$), on the other hand, is calculated using the ideal gas law:

\begin{equation}
    \rho_g = \frac{P_{bi}M_g}{RT},
\end{equation}

\noindent here, $M_g$ refers to the molecular weight of air, while $R$ denotes the universal gas constant and $T$ represents the temperature of the surrounding environment. The total outlet flow rate can be determined using the following relationship:

\begin{equation}
    w_{total} = w_{g,out} + w_{l,out} = \theta_{top}\sqrt{\rho_{mix}(P_{rh}-P_{atm})},
\end{equation}

\noindent $P_{atm}$ represents the atmospheric pressure, while $\theta_{top}$ denotes the flow coefficient of the top valve. Additionally, we assume that the proportion between the liquid and total outlet flow rates remains consistent with the liquid fraction (represented by $a_l$) present in the mixture. This assumption can be expressed as follows:

\begin{equation}
    \alpha_l = \frac{m_{l}}{m_{total}} = \frac{w_{l,out}}{w_{total}}.
\end{equation}

\subsection{Offline Digital Twin Identification}

The digital twin identification offline step of this methodology is a crucial phase in developing a digital twin framework, which aims to address the challenges associated with robustness, uncertainty, and the integration of various learning techniques for optimal and autonomous decision-making in gas-lift processes. 
Overall, this step of offline training involves identifying the AI models using computationally intensive data in an offline environment. The objective is to create accurate and reliable models based on historical data, which can then be deployed in an online environment for real-time decision-making.

Data acquisition is the first step in constructing the gas lift offline digital twin. The data quality and quantity are fundamental for adequately representing the process domain. Synthetic data was generated through the previously validated phenomenological model. As discussed in Section \ref{Design_of_Experiments}, we used the DoE methodology to plan data acquisitions. Therefore, LHS was applied to generate 4000 experiments for input data. 

Each experiment consists of a given input that was applied to the process for some time necessary for the system to reach a steady state, defined as 100 seconds. Hence, the database is constituted of 400,000 points. It is essential to highlight the complete transient responses of each experiment were stored for DT identification.

To choose the input variables for the Digital Twin model, a Gram-Schmidt orthogonalization method was employed to analyze the impact of operational variables on the flow rates of water and air injected into the system. This variable ranking methodology, developed by \cite{Nogueira2016}, enables identifying which process variables have the most significant impact on the outcome of the process. For further details, please refer to \cite{Nogueira2016}.

Hence, the orthogonalization analysis pointed out that the gas flow rates for each corresponding well (FI104, FI105, and FI106) and the pump outlet pressure are the variables with the greatest impact on the water and gas flow rates produced in the gas lift system. Therefore, these are the variables used as input in the data-driven model. The variables selected as inputs to the data-driven model, as well as the defined limits for generating input data, are presented in Table \ref{TAB:Inputs_limits}. The limits were established according to the operational conditions of the real plant.

\begin{table}[width=.8\linewidth,pos=h!]
\caption{Operating conditions bounds given to the LHS design of experiments.}
\begin{tabular*}{\tblwidth}{@{} Lcccc@{} }
\toprule
  & $Q_{g,1}/ (sLmin^{-1})$ & $Q_{g,2}/ (sLmin^{-1})$ & $Q_{g,3}/ (sLmin^{-1})$ & $P_{pump}/ (bar)$                \\ \midrule
Minimum & 1                    & 1                    & 1                    & 1.3                  \\
Maximum & 5                    & 5                    & 5                    & 4                    \\
\bottomrule
\end{tabular*}
\label{TAB:Inputs_limits}
\end{table}

It is worth noting that \cite{Matias2022} also uses these variables (Table \ref{TAB:Inputs_limits}) as input in the system for real-time optimization and process control. On the other hand, the reservoir valve openings (CV101, CV102, and CV103) are used to introduce unmeasured disturbances in the system.

\begin{figure}[h!]
	\centering
		\includegraphics[scale=.5]{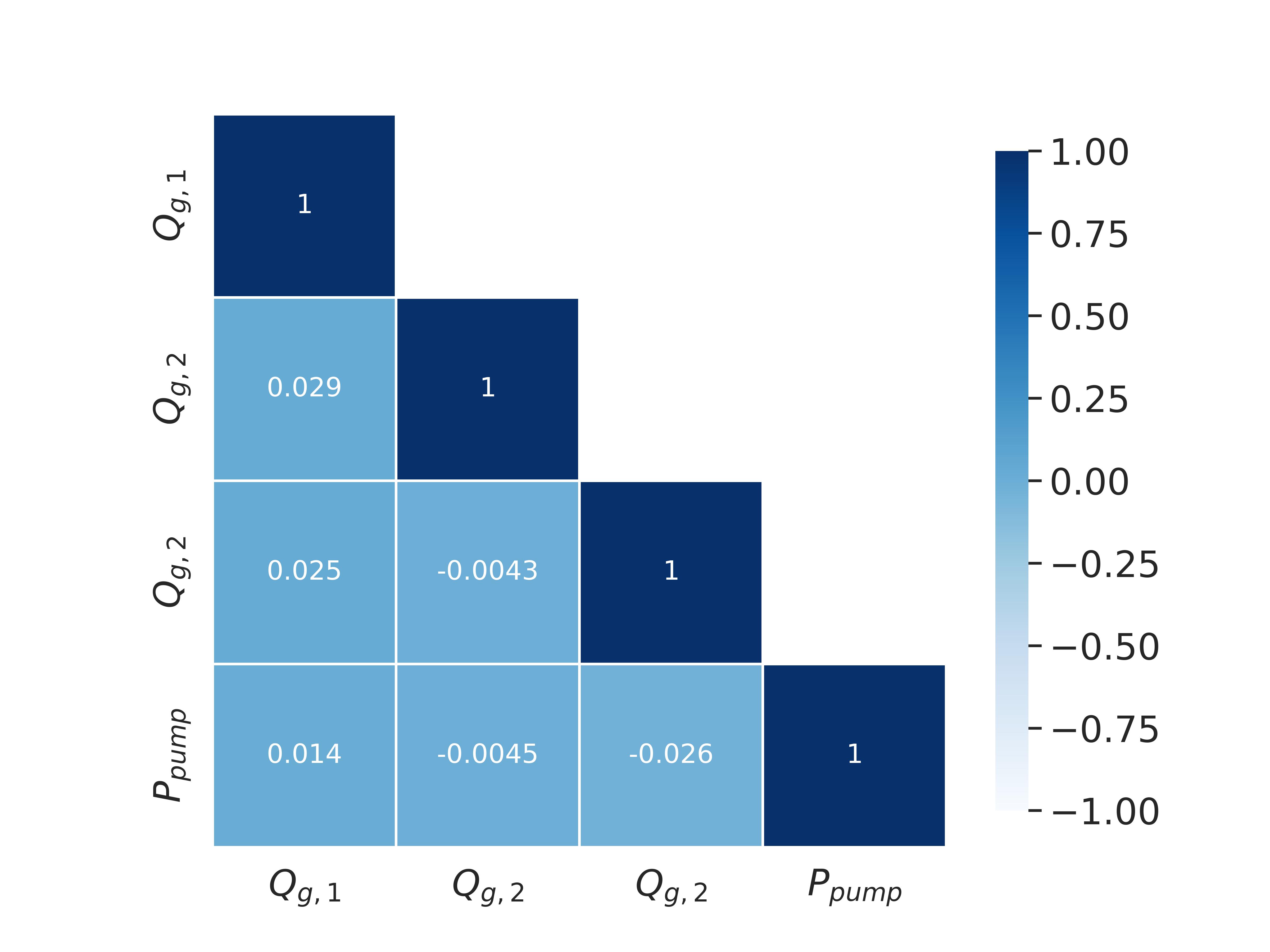}
    \caption{Correlation heatmap of the LHS inputs signals.}
    \label{Correlation}
\end{figure}

After generating the input matrix using the LHS method, a correlation evaluation between the variables was performed. Figure \ref{Correlation} shows a heat map of all input variables of the model. The results indicate that the correlations have low values close to zero. This is an essential factor suggesting that the input space provided by LHS was well designed, which avoids data skew and, consequently, the training will not present undesirable biases.

\begin{figure}[h!]
	\centering
		\includegraphics[scale=.6]{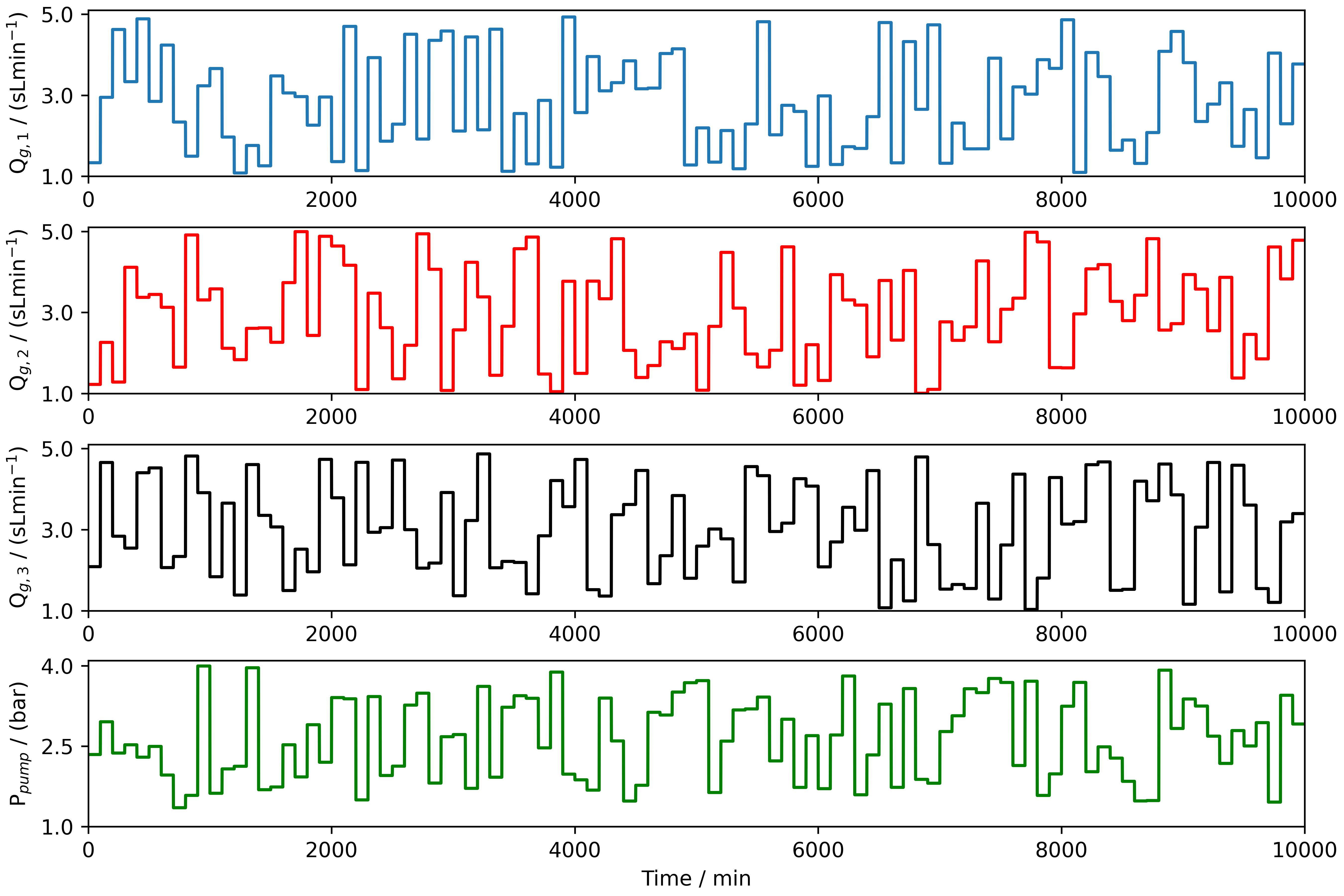}
    \caption{Experimental input sequence.}
    \label{Input_LHS}
\end{figure}

The limits for the design of experiments of the input were adjusted to their respective engineering dimensions, presented in Table \ref{TAB:Inputs_limits}. Subsequently, these perturbations were inserted into the phenomenological model to generate data sets that comprised the output matrix of the model. Figure \ref{Input_LHS} illustrates the input representation used to induce disturbances in the phenomenological model.

After constructing the output matrix of the phenomenological model, it was necessary to define the appropriate data structure for the AI model. As discussed in Section \ref{Predictor_structure_identification}, we chose to use a model known as Nonlinear Autoregressive Network with Exogenous Inputs (NARX) to predict nonlinear behaviors in dynamic systems of the gas lift model. This prediction model allows the introduction of exogenous variables. However, it was necessary to define the predictors' embedding dimensions, i.e., the ideal number of previous inputs $(N_a)$, and previous outputs $(N_b)$, to organize the training, validation, and test datasets.

\begin{figure}[h!]
	\centering
		\includegraphics[scale=.035]{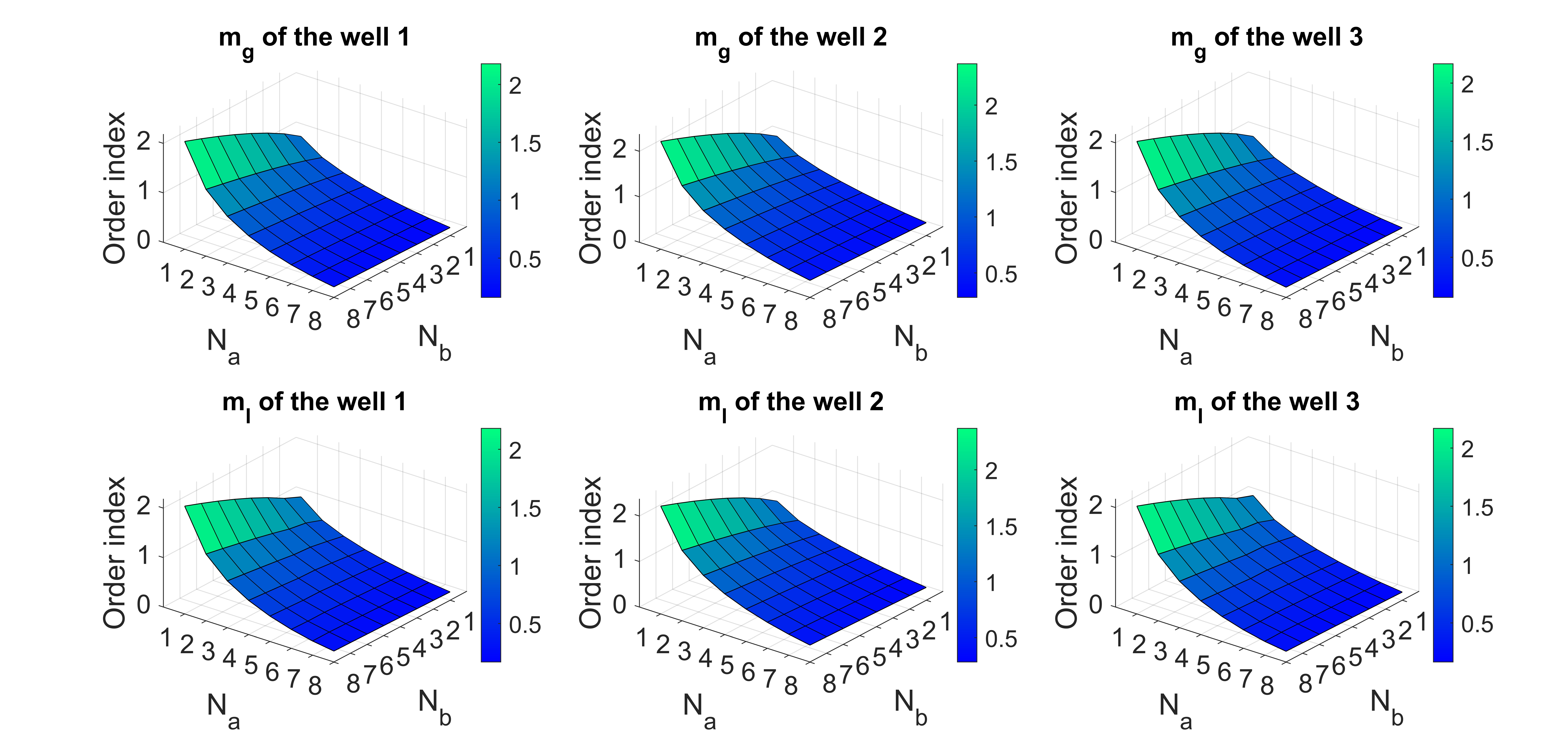}
    \caption{Lipschitz coefficients results for the predictor embedding dimensions.}
    \label{Lipischit}
\end{figure}

Figure \ref{Lipischit} presents the Lipschitz coefficients to determine the optimal number of delays for the inputs and outputs in a Nonlinear Autoregressive with exogenous inputs predictor. The graph shows the relationship between the number of delays and the Lipschitz Index, which measures the predictability of the system.

The slope of the Lipschitz index is used to determine the optimal number of delays, with the slope tending to zero at the optimal point. In the present case, the optimal number of delays for the inputs and outputs can be seen on the graph as the points where the slope of the Lipschitz index approaches zero.

Figure \ref{Lipischit} concisely represents the relationship between the number of delays and the Lipschitz index, making it a useful tool for determining the optimal number of delays in NARX predictors. As seen in the figure, the optimal values for the number of delays in all the inputs and outputs are five and two, respectively. These values are critical for optimizing the predictability of the NARX predictor and ensuring that the model accurately captures the system's behavior.

The HYPERBAND method is a powerful optimization technique that was applied to identify the optimal hyperparameters for an AI model used for uncertainty identification. The methodology, described in Section \ref{Hyperparameters}, uses a combination of random sampling and bracketing to efficiently explore the hyperparameter space and determine the best values for each hyperparameter. In this study, two groups of hyperparameters were considered: optimization parameters (learning rate and mini-batch size) and structural parameters (number of neurons, activation function, and number of the dense layer).

To ensure that the model was suitable for prediction purposes, the type of layer was fixed as a simple feedforward layer. This decision was based on the recommendations of \cite{Rebello2022}, who present a comprehensive guide for selecting Neural Network structures for prediction and simulation purposes. The search limits for each hyperparameter were defined as described in Table \ref{TAB:Hyperparameters}. The results of the HYPERBAND search are presented in Table \ref{result_hyperband}. 

\begin{table}[width=.9\linewidth,cols=4,pos=h]
\renewcommand{\arraystretch}{1.3} 
\caption{Results of best hyperparameters for each performance indicator for NN.}
\begin{tabular*}{\tblwidth}{@{} LLLL@{} }
\toprule
Wells & Hyperparameters & m$_{g}$  & m$_{l}$ \\
\midrule
\multirow{5}{*}{Well 1} 
& Initial learning rate		& $1\times 10^{-3}$	& $1\times 10^{-3}$ 	\\
& Number of dense layers		& 2	& 2   \\
& Activation function in each layer 		& \{relu, linear\}	& \{relu, linear\}   \\
& Number of neurons in each layer     		& \{60, 1\}	& \{60, 1\}   \\
& Number of parameters for each layer        		& \{900, 61\}	& \{900, 61\}   \\
\hline
 \multirow{5}{*}{Well 2} 
& Initial learning rate		& $1\times 10^{-3}$	& $1\times 10^{-3}$ 	\\
& Number of dense layers		& 2	& 2   \\
& Activation function in each layer 		& \{relu, linear\}	& \{relu, linear\}   \\
& Number of neurons in each layer     		& \{40, 1\}	& \{70, 1\}   \\
& Number of parameters for each layer      & \{600, 41\}	& \{1050, 71\}   \\
\hline
 \multirow{5}{*}{Well 3} 
& Initial learning rate		& $1\times 10^{-3}$	& $1\times 10^{-3}$ 	\\
& Number of dense layers		& 2	& 2   \\
& Activation function in each layer 		& \{relu, linear\}	& \{relu, linear\}   \\
& Number of neurons in each layer     		& \{60, 1\}	& \{60, 1\}   \\
& Number of parameters for each layer      & \{900, 61\}	& \{900, 61\}   \\
 \bottomrule
\end{tabular*}
\label{result_hyperband}
\end{table}

\begin{figure}[h!]
	\centering
    \begin{subfigure}{0.3\textwidth}
        \centering
		\includegraphics[scale=.45]{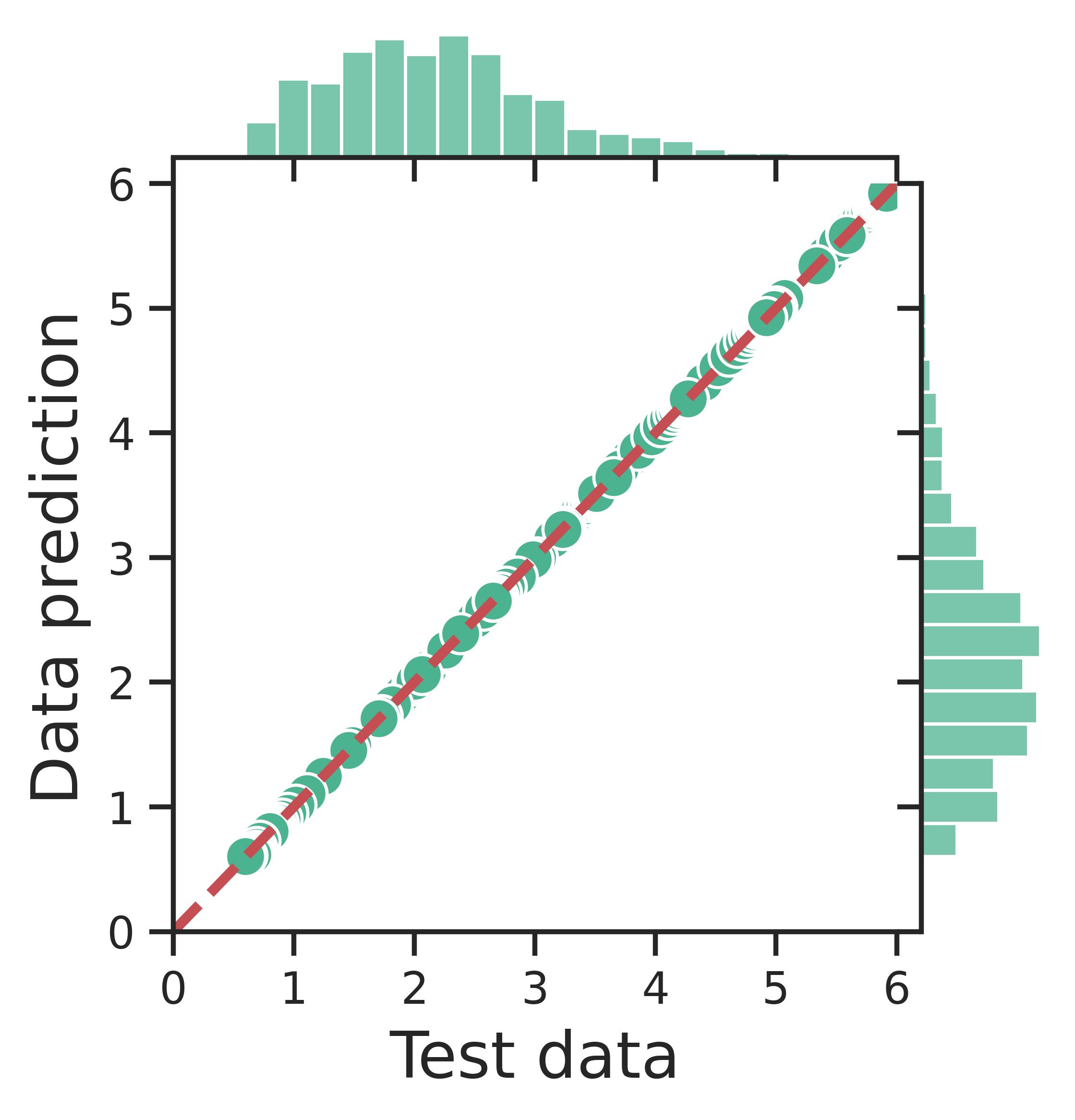}
	\caption{}

\end{subfigure}
    \centering
\begin{subfigure}{0.3\textwidth}
	\centering
		\includegraphics[scale=.45]{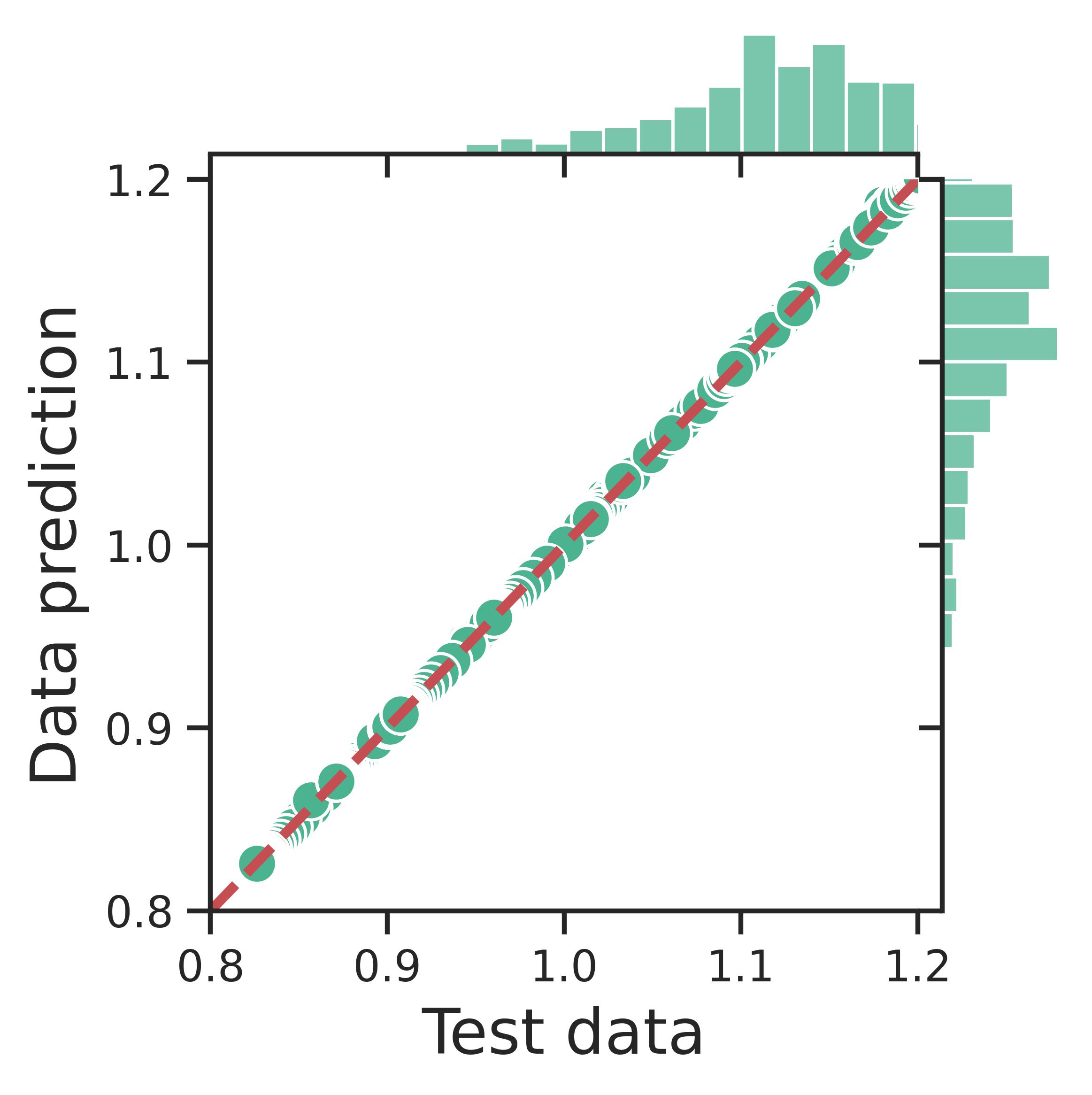}
	\caption{}
	
\end{subfigure}
	
 \begin{subfigure}{0.3\textwidth}
	\centering
		\includegraphics[scale=.45]{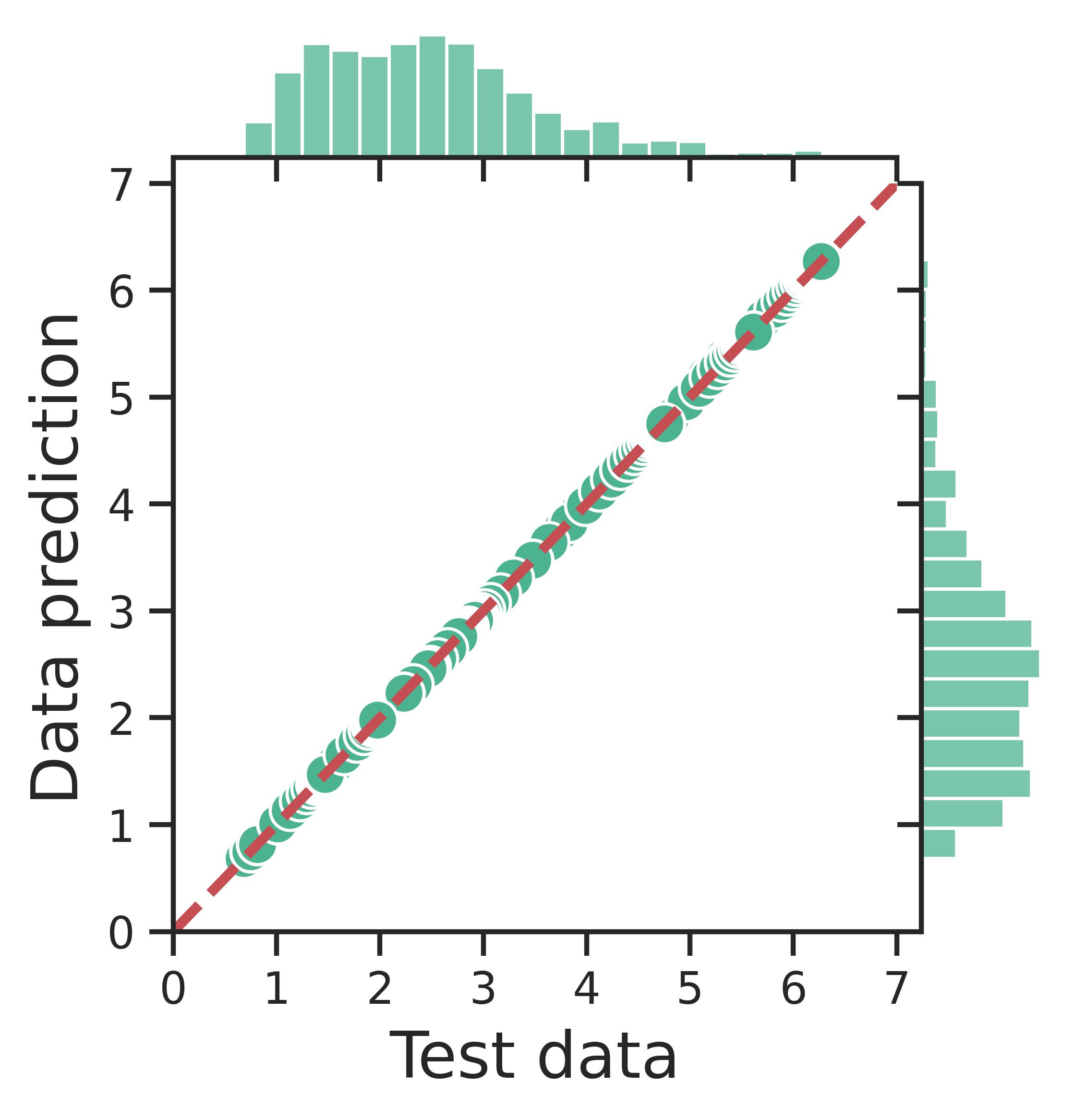}
	\caption{}
	
\end{subfigure}
	
  \begin{subfigure}{0.3\textwidth}
	\centering
		\includegraphics[scale=.45]{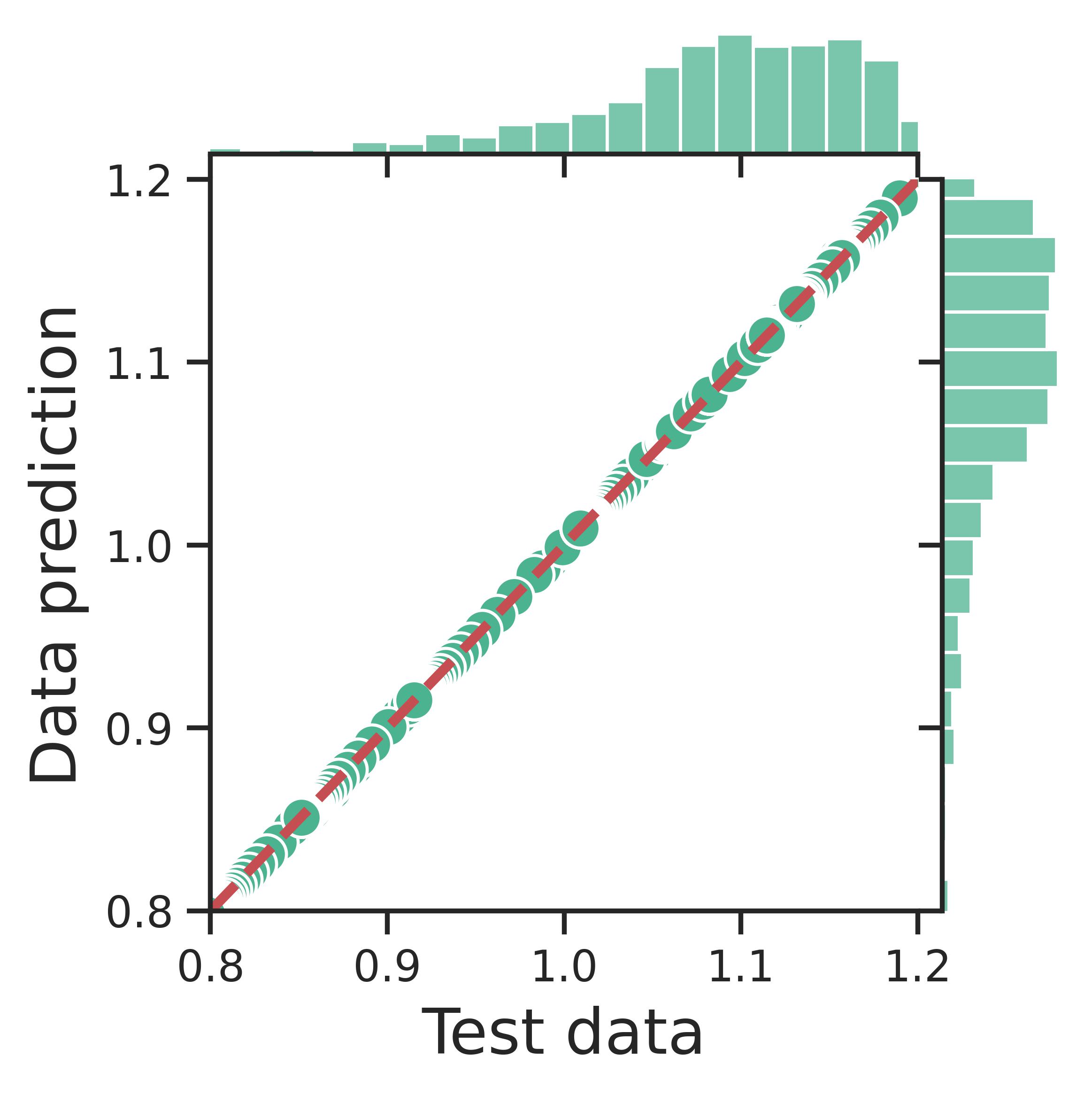}
	\caption{}
	
\end{subfigure}
  \begin{subfigure}{0.3\textwidth}
	\centering
		\includegraphics[scale=.45]{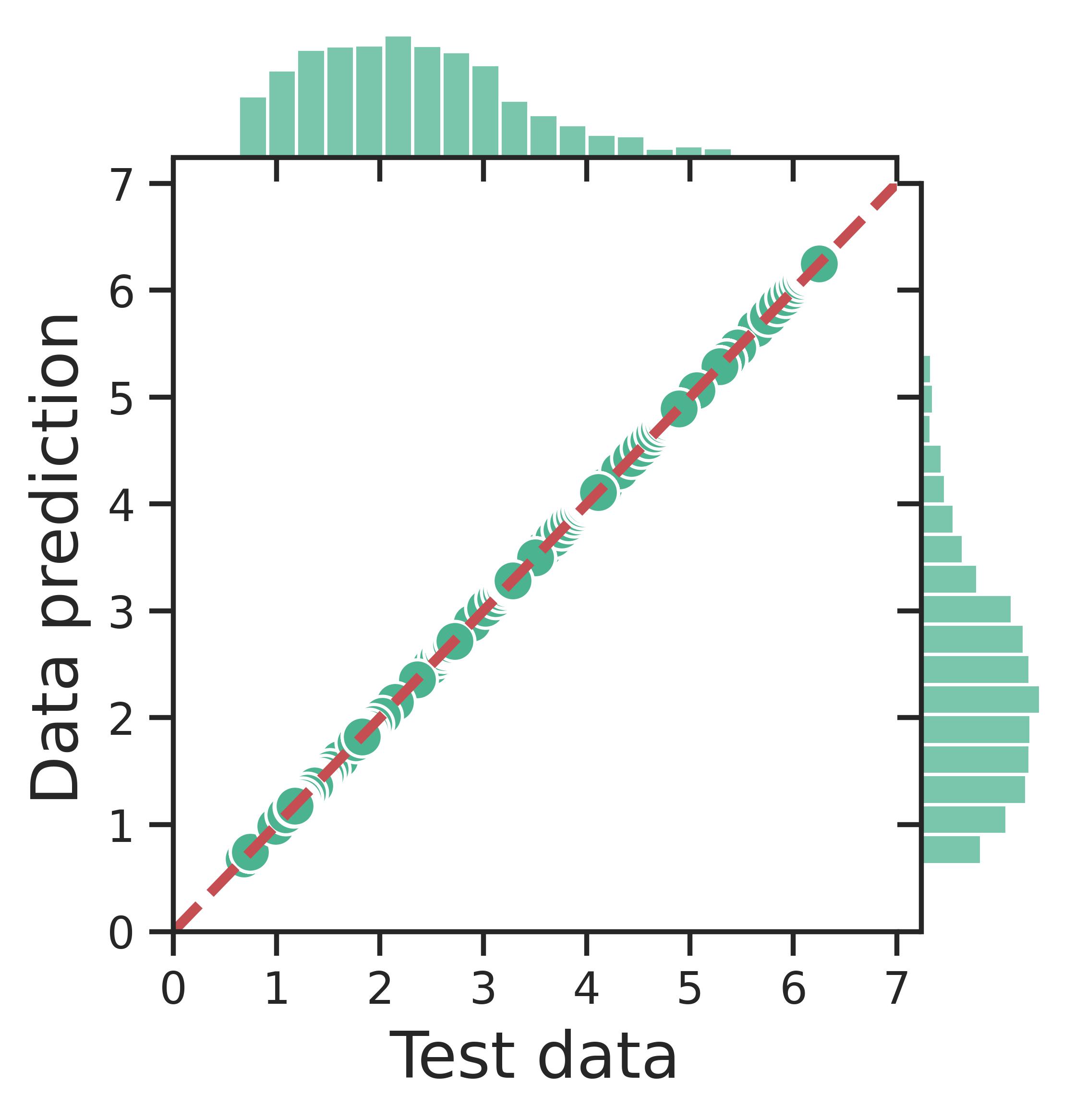}
	\caption{}
	
\end{subfigure}
  \begin{subfigure}{0.3\textwidth}
	\centering
		\includegraphics[scale=.45]{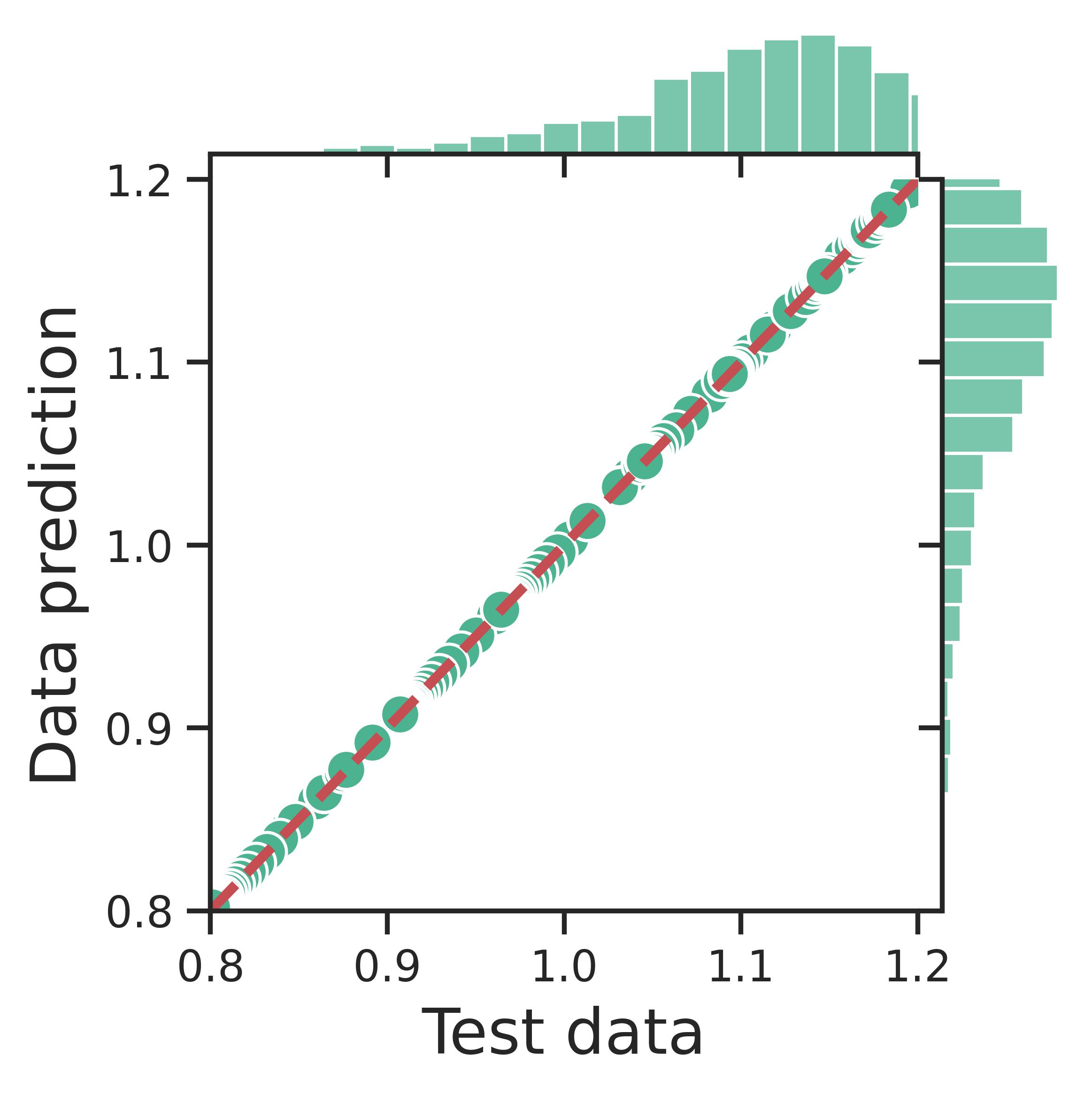}
	\caption{}
	
\end{subfigure}
	
    \caption{Parity graphics of the model test, (a) $m_g$ of the well 1, (b) $m_g$ of the well 2, (c) $m_g$ of the well 3, (d) $m_l$ of the well 1, (e) $m_l$ of the well 2, (f) $m_l$ of the well 3}
    \label{paridade}
\end{figure}

Figure \ref{paridade} presents a crucial aspect of evaluating the performance of the AI model that was identified as the base for the digital twin. The parity graph visually compares the model's predictions and the actual test data. This graph provides a clear insight into the model's behavior and its ability to capture the underlying system's dynamics accurately.

The random distribution of the points along the diagonal line in the graph indicates the model's accuracy. The parity graph shows that the AI model's predictions align well with the test data across the complete range of validation, and the residuals are randomly distributed. This is an important verification of the optimization procedure's effectiveness in finding the best parameters for the AI model.

Statistically, random residuals indicate that the optimization procedure has reached a satisfactory result in the parameter identification process. In other words, the AI model has learned the underlying patterns and relationships between the inputs and outputs and can accurately make predictions. 

\begin{figure}[h!]
	\centering
    \begin{subfigure}{0.4\textwidth}
		\includegraphics[scale=.5]{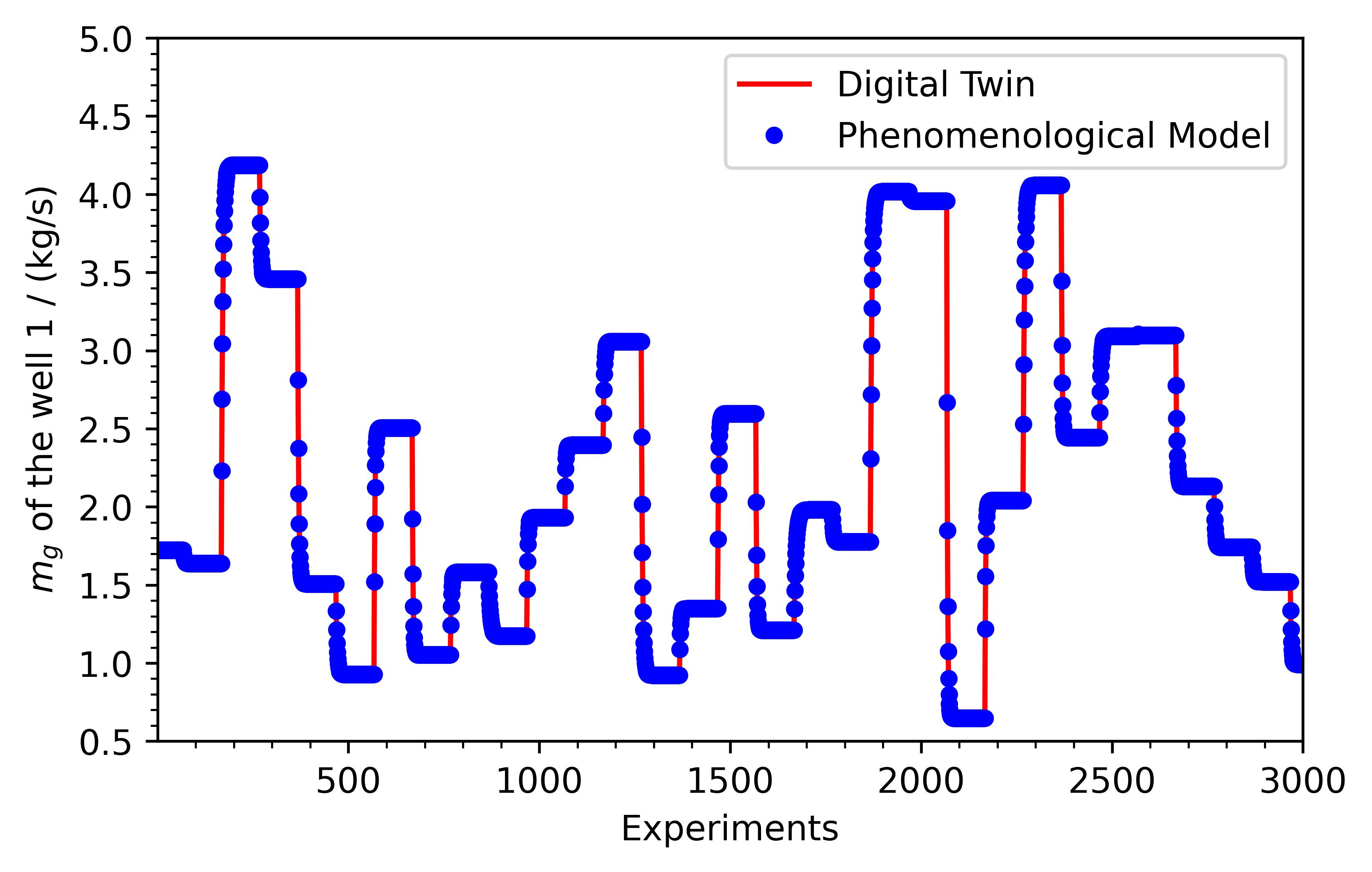}
	\caption{}
	
\end{subfigure}
    \centering
\begin{subfigure}{0.4\textwidth}
	\centering
		\includegraphics[scale=.5]{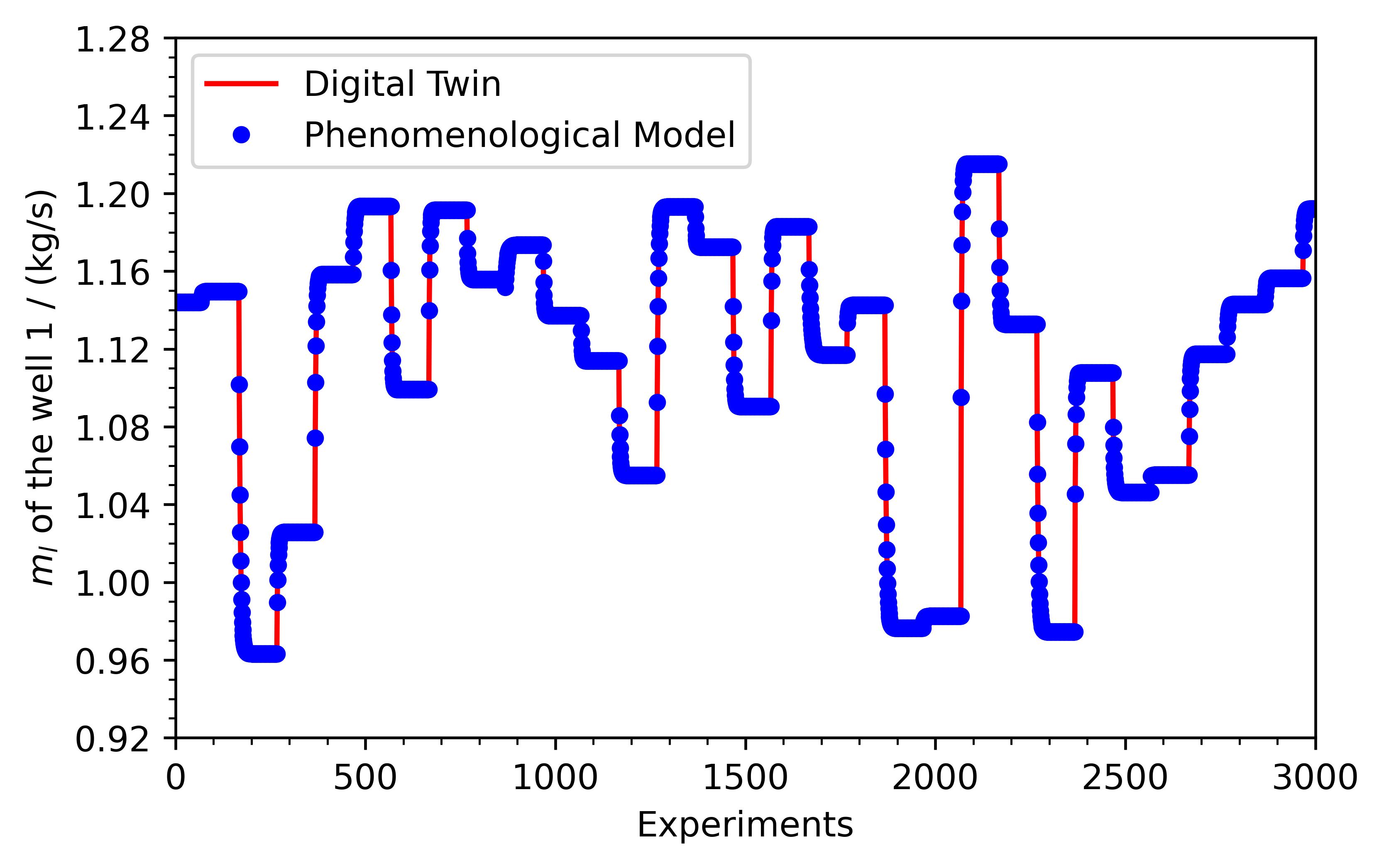}
	\caption{}
	
\end{subfigure}
	
 \begin{subfigure}{0.4\textwidth}
	\centering
		\includegraphics[scale=.5]{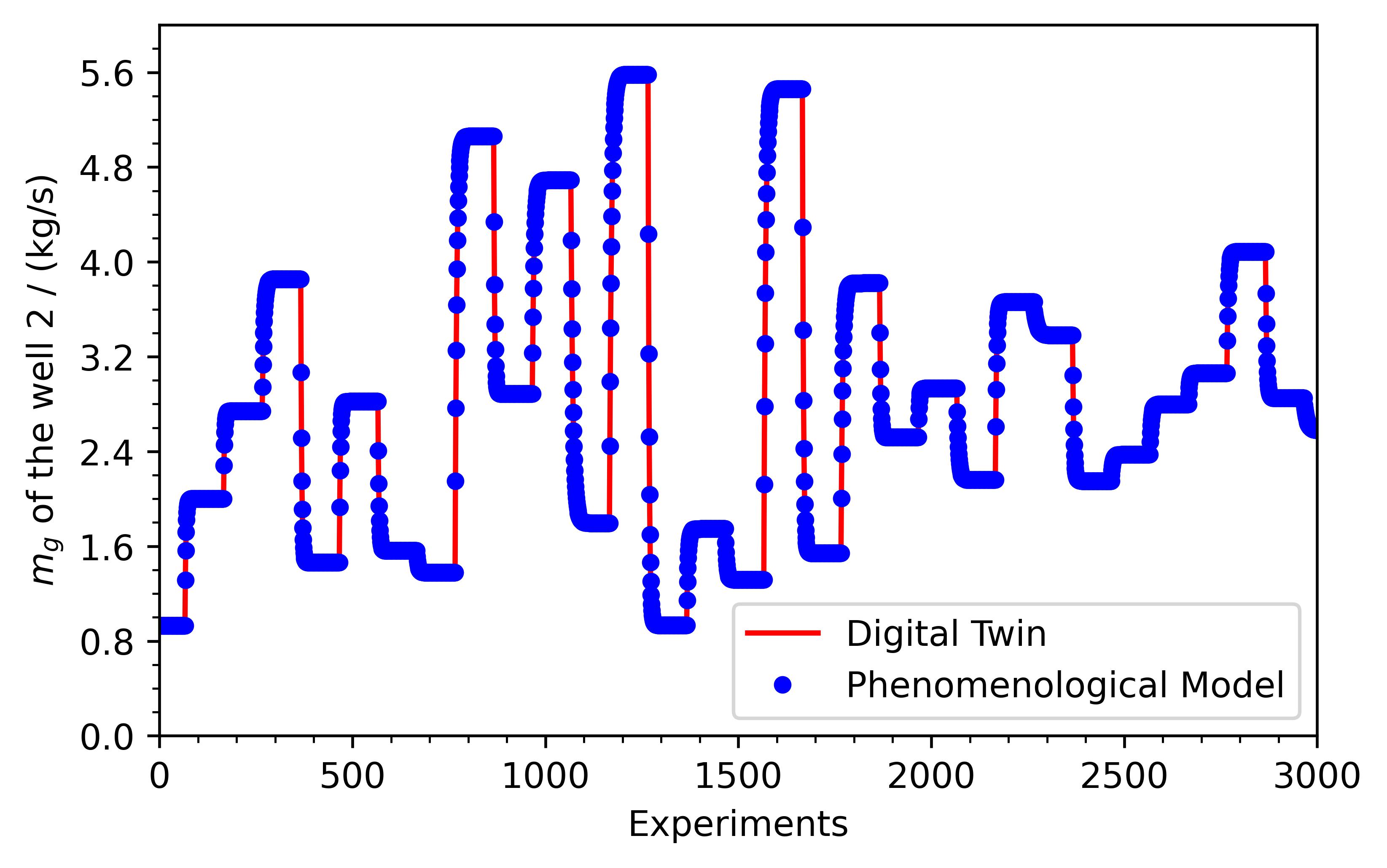}
	\caption{}
	
\end{subfigure}
	
  \begin{subfigure}{0.4\textwidth}
	\centering
		\includegraphics[scale=.5]{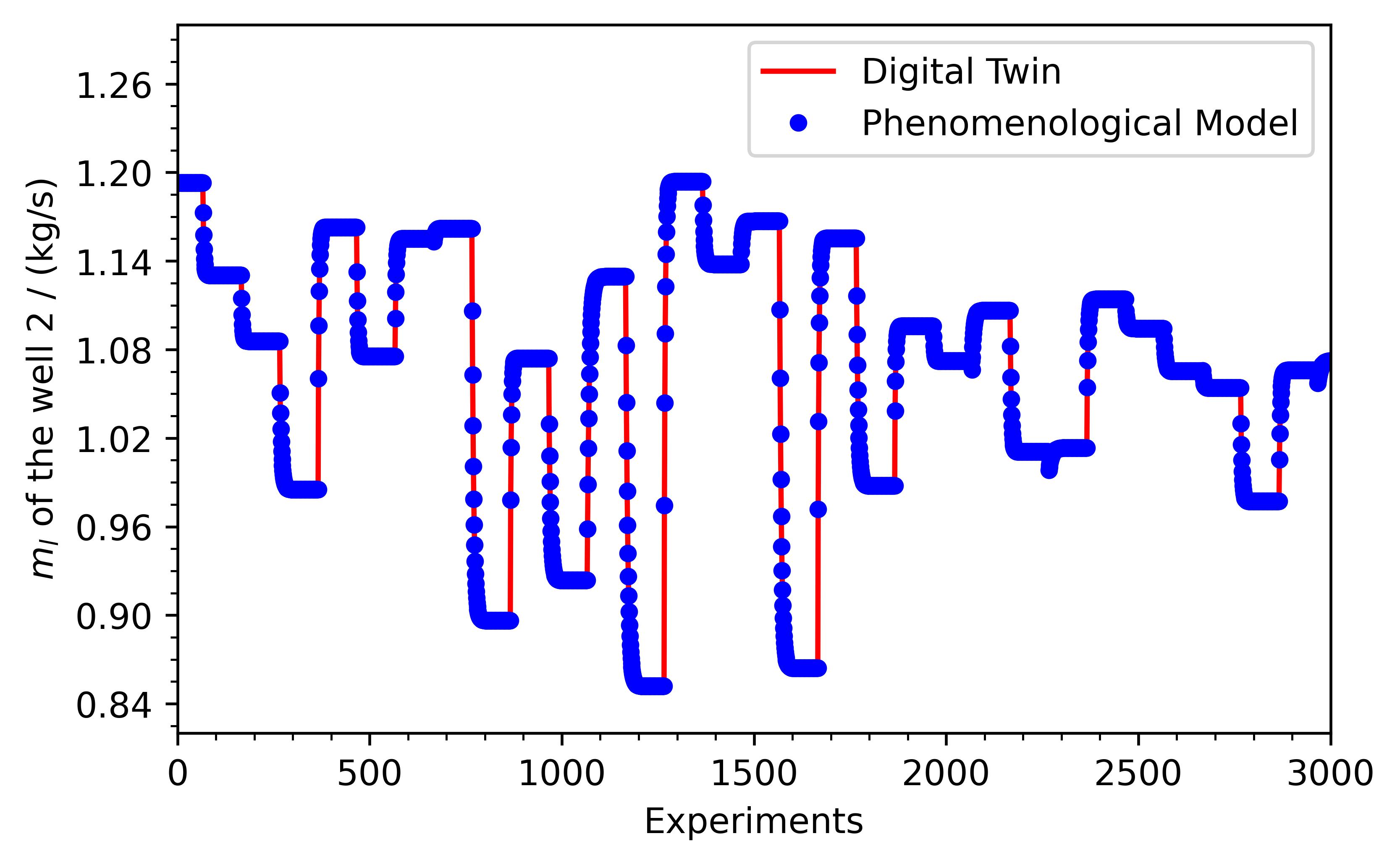}
	\caption{}
	
\end{subfigure}
  \begin{subfigure}{0.4\textwidth}
	\centering
		\includegraphics[scale=.48]{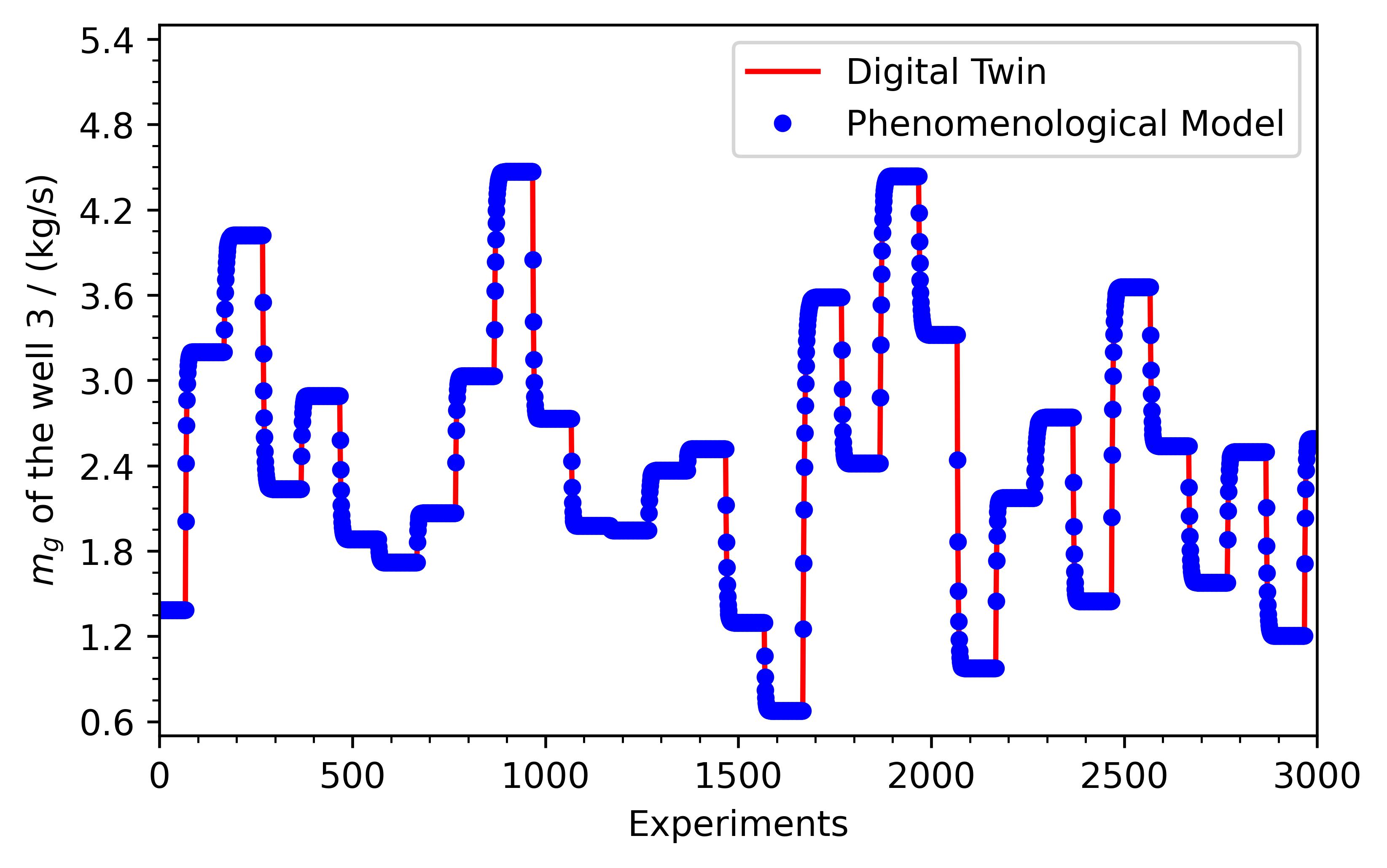}
	\caption{}
	
\end{subfigure}
  \begin{subfigure}{0.4\textwidth}
	\centering
		\includegraphics[scale=.5]{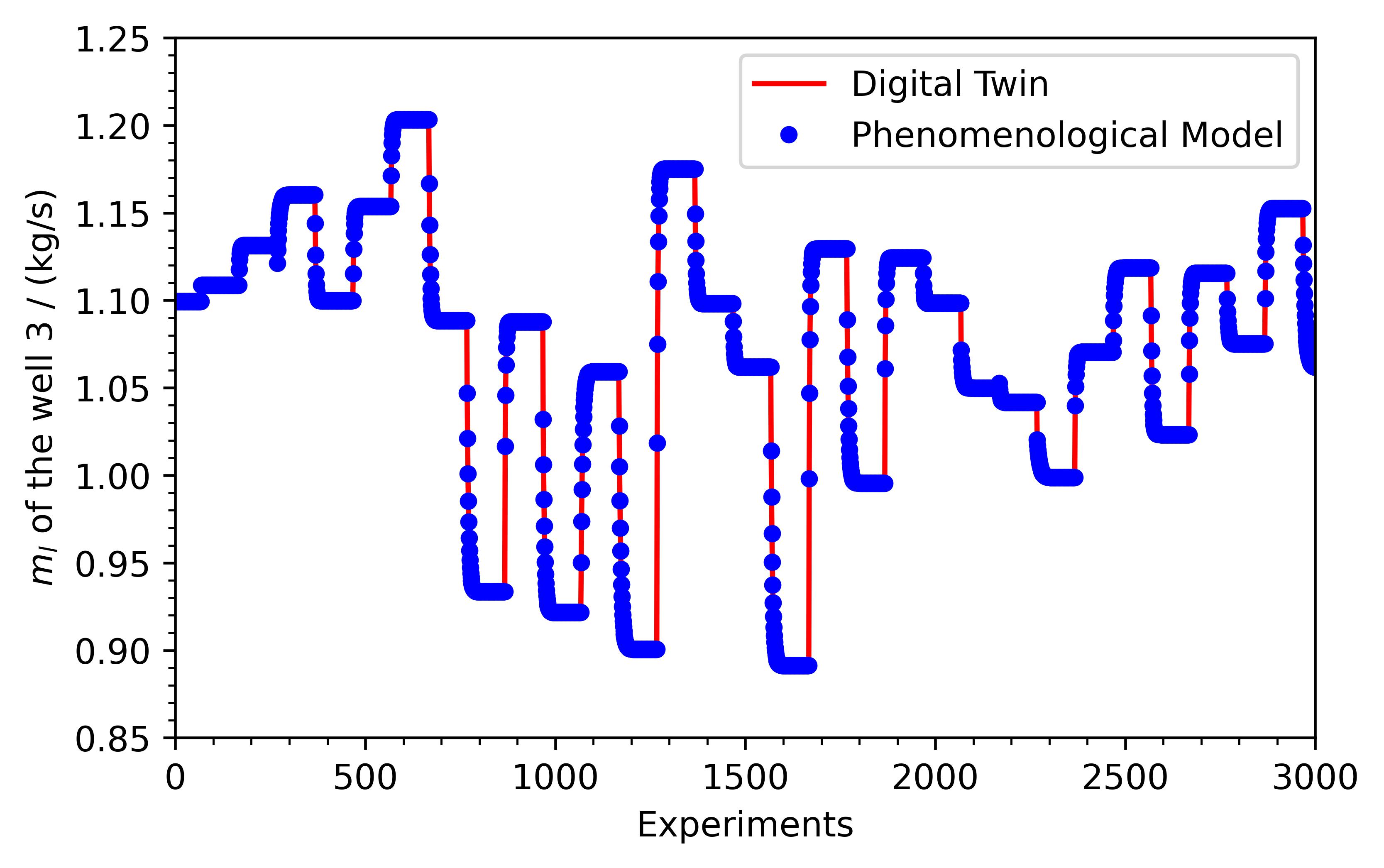}
	\caption{}
	
\end{subfigure}
	
    \caption{Neural network models test for each process output}
    \label{test_dinamic}
\end{figure}

In line with the parity analysis, the effectiveness of the AI model is highlighted in Figure \ref{test_dinamic}, where the model's prediction is compared with the test data over time. The graph represents the model's behavior and ability to track the test data dynamics accurately.

The results of the AI model performance are presented in Table \ref{MSE_MAE}, showcasing the accuracy of the model's predictions. The metrics used to evaluate the models' performance include the Mean Absolute Error (MAE) and Mean Squared Error (MSE), both of which are commonly used measures of the difference between the actual and predicted values. The results show that all of the models have low MAE and MSE values for the test data, indicating that the models have been successfully identified and are making accurate predictions. The low error values highlight the reliability of the models and their ability to precisely track the dynamics of the test data over time. Overall, the results presented in Table \ref{MSE_MAE} provide evidence of the success of the AI model identification process.

\begin{table}[width=.7\linewidth,cols=4,pos=h]
\renewcommand{\arraystretch}{1.3} 
\caption{Models' performance indicators of the validation.}
\begin{tabular*}{\tblwidth}{@{} LLLL@{} }
\toprule
Wells & Metrics & m$_{g}$ & m$_{l}$\\
\midrule
\multirow{2}{*}{Well 1} 
& MSE		& $1.37\times 10^{-7}$	& $1.62\times 10^{-7}$ 	\\
& MAE		& $2.69\times 10^{-4}$	& $2.72\times 10^{-4}$   \\

 \multirow{2}{*}{Well 2} 
& MSE		& $8.46\times 10^{-7}$	& $1.55\times 10^{-7}$ 	\\
& MAE		& $9.79\times 10^{-4}$	& $2.96\times 10^{-4}$   \\

 \multirow{2}{*}{Well 3} 
& MSE		& $1.23\times 10^{-7}$	& $1.65\times 10^{-7}$ 	\\
& MAE		& $2.80\times 10^{-4}$	& $2.68\times 10^{-4}$   \\
 \bottomrule
\end{tabular*}
\label{MSE_MAE}
\end{table}

The next step in the proposed methodology is to assess the uncertainty of the digital twin-base model. The objective of this step is to improve the robustness and reliability of the AI model, making it an even more effective digital twin for the underlying system. To achieve this, the methodology outlined in Section \ref{incerteza} was followed.

\begin{figure}
\begin{minipage}[t]{\linewidth}
  \begin{subfigure}{\linewidth}
  \centering
  \includegraphics[width=.5\linewidth]{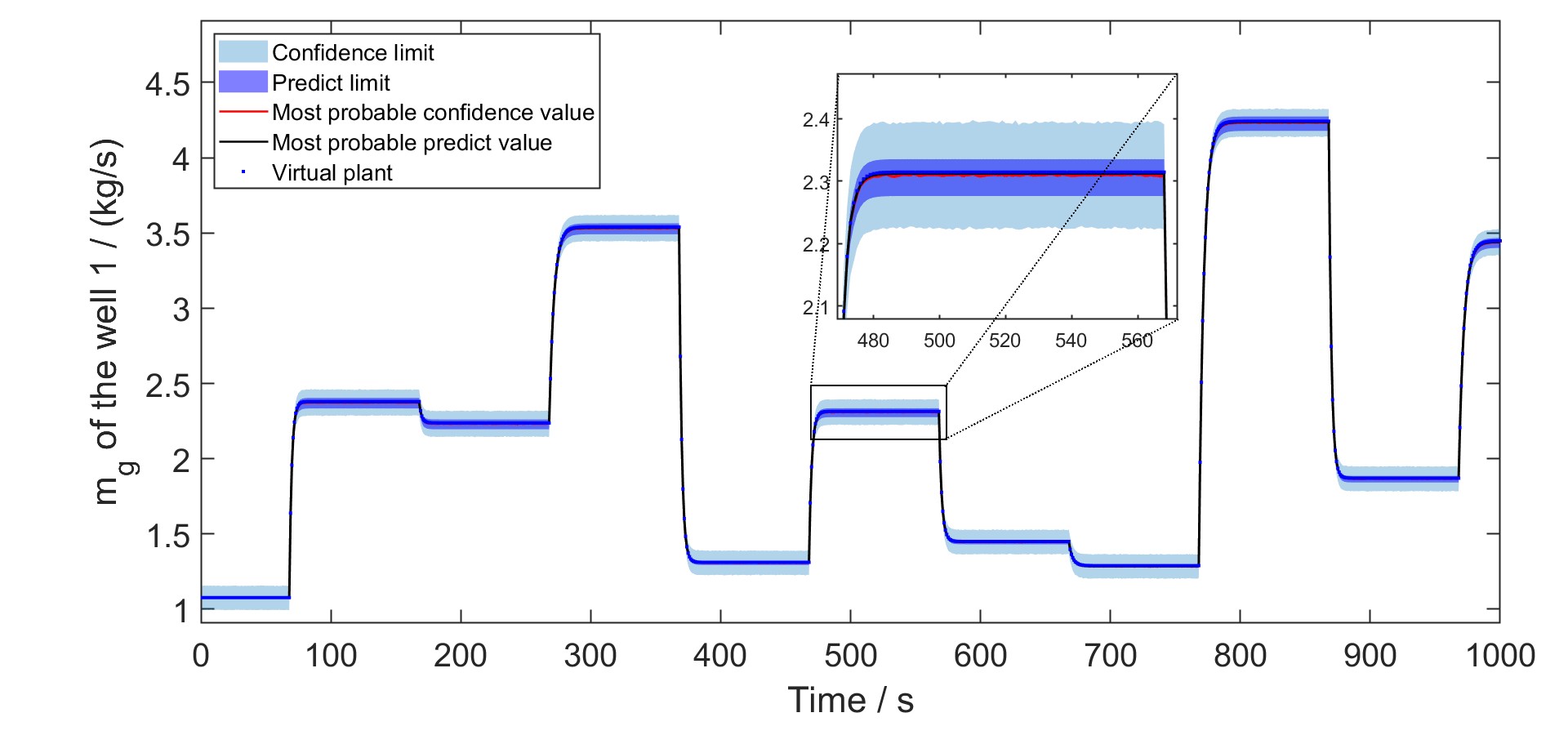}
  \caption{$m_g$ well 1}
  \end{subfigure}\par\bigskip
  \begin{subfigure}{\linewidth}
  \centering
  \includegraphics[width=.5\linewidth]{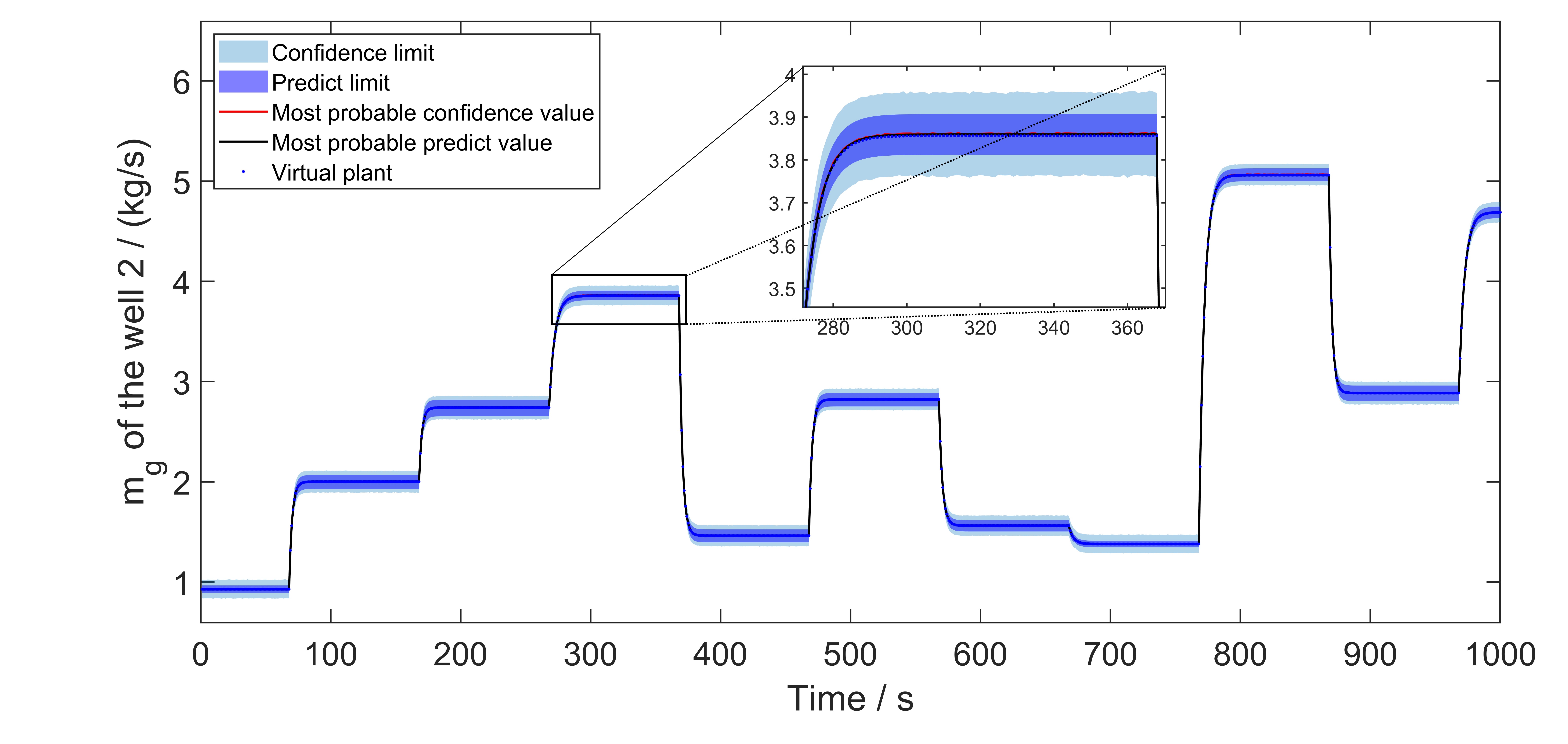}
  \caption{$m_g$ well 2}
  \end{subfigure}\par\bigskip
  \begin{subfigure}{\linewidth}
  \centering
  \includegraphics[width=.5\linewidth]{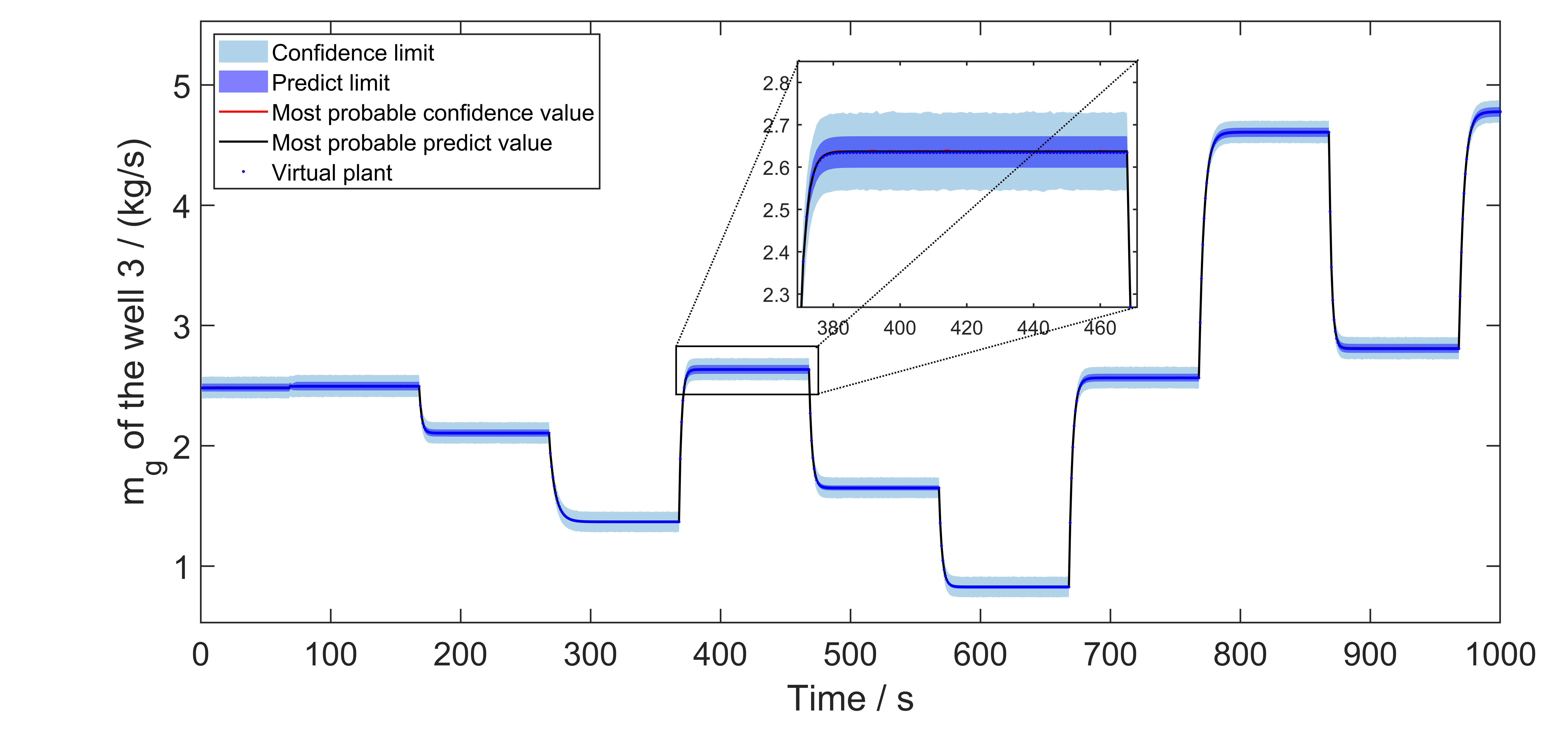}
  \caption{$m_g$ well 3}
  \end{subfigure}
\end{minipage}
\caption{Prediction uncertainty of neural network models sampled for test data}
\label{Figura_parte_1}
\end{figure}

\begin{figure}
\begin{minipage}[t]{\linewidth}
  \begin{subfigure}{\linewidth}
  \centering
  \includegraphics[width=.5\linewidth]{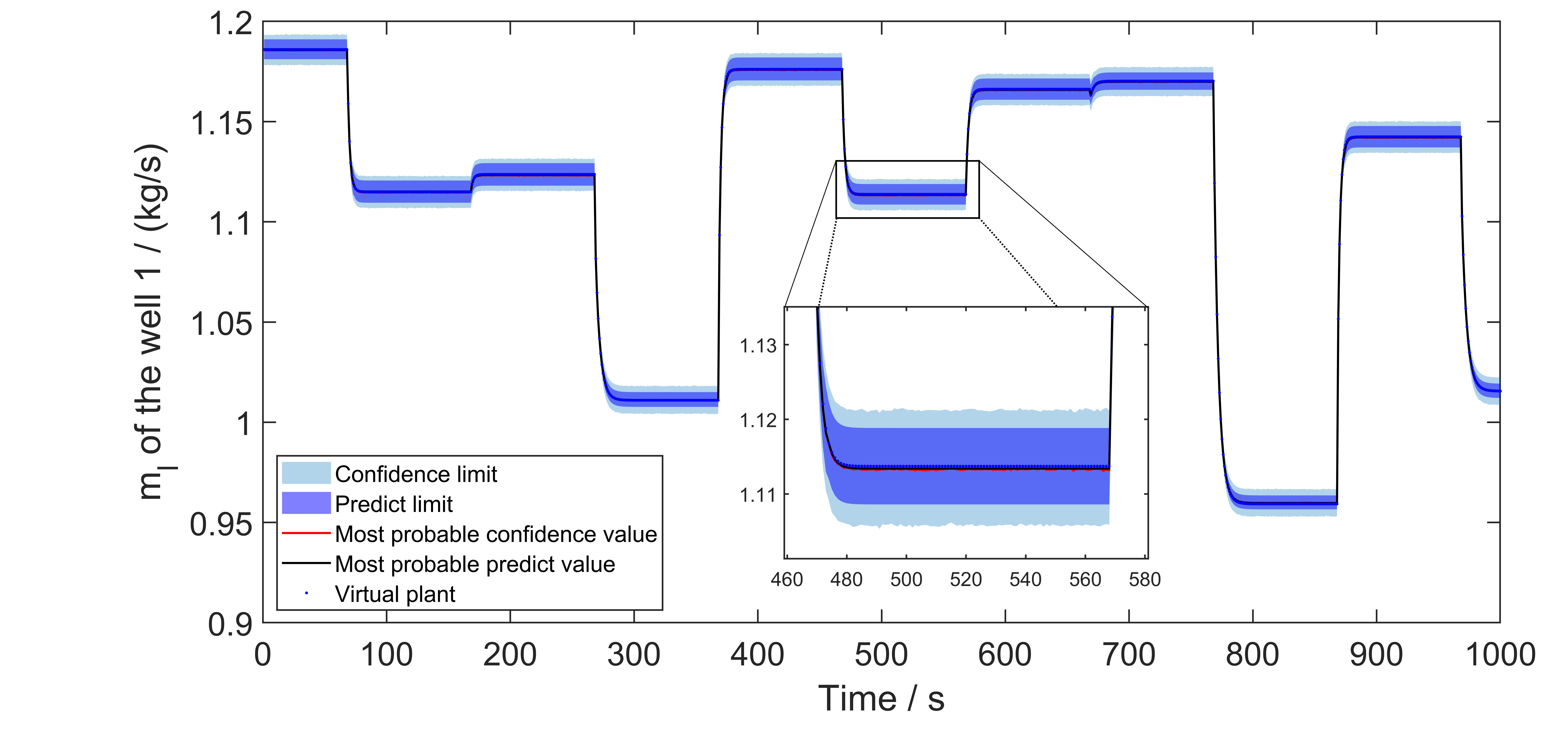}
  \caption{$m_l$ well 1}
  \end{subfigure}\par\bigskip
  \begin{subfigure}{\linewidth}
  \centering
  \includegraphics[width=.5\linewidth]{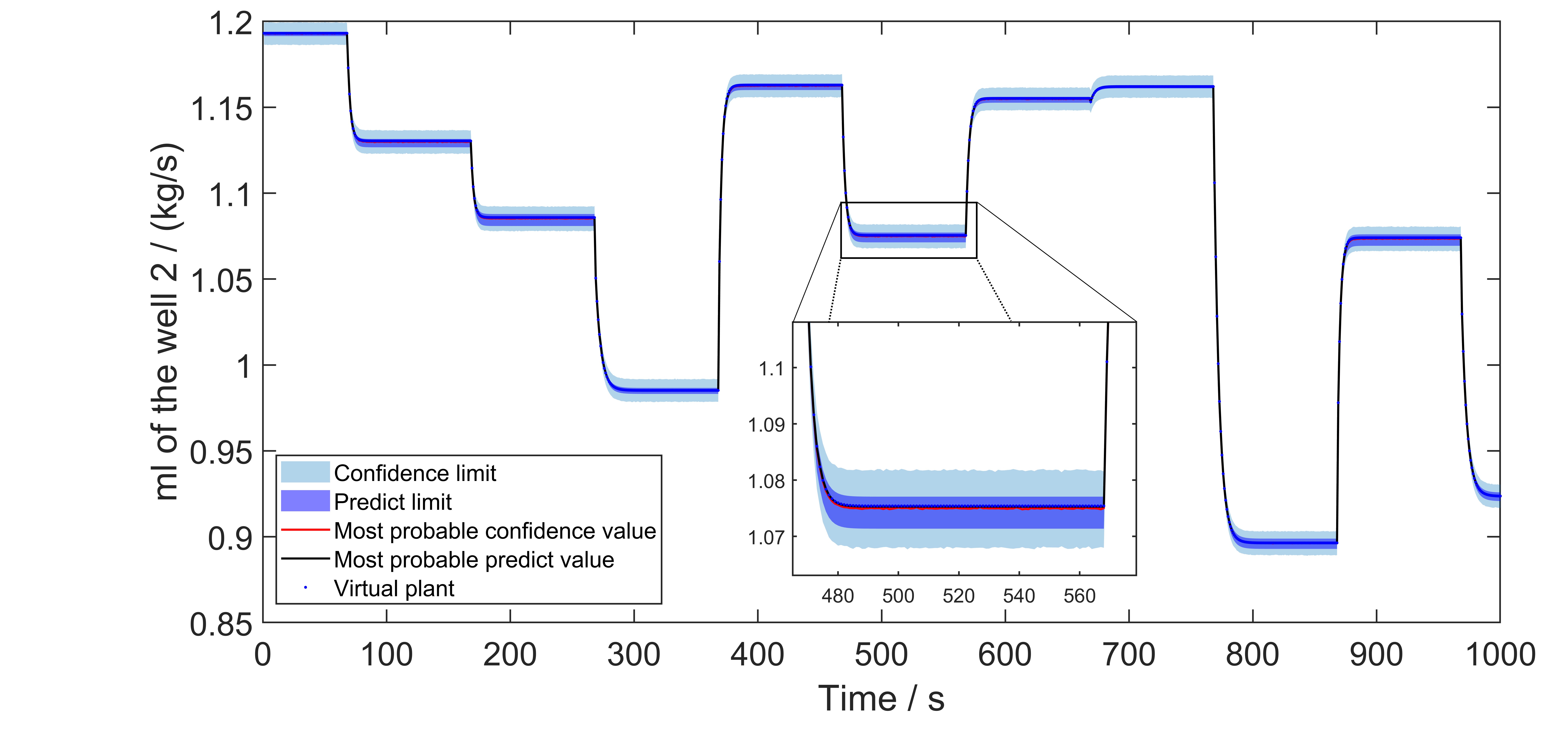}
  \caption{$m_l$ well 2}
  \end{subfigure}\par\bigskip
  \begin{subfigure}{\linewidth}
  \centering
  \includegraphics[width=.5\linewidth]{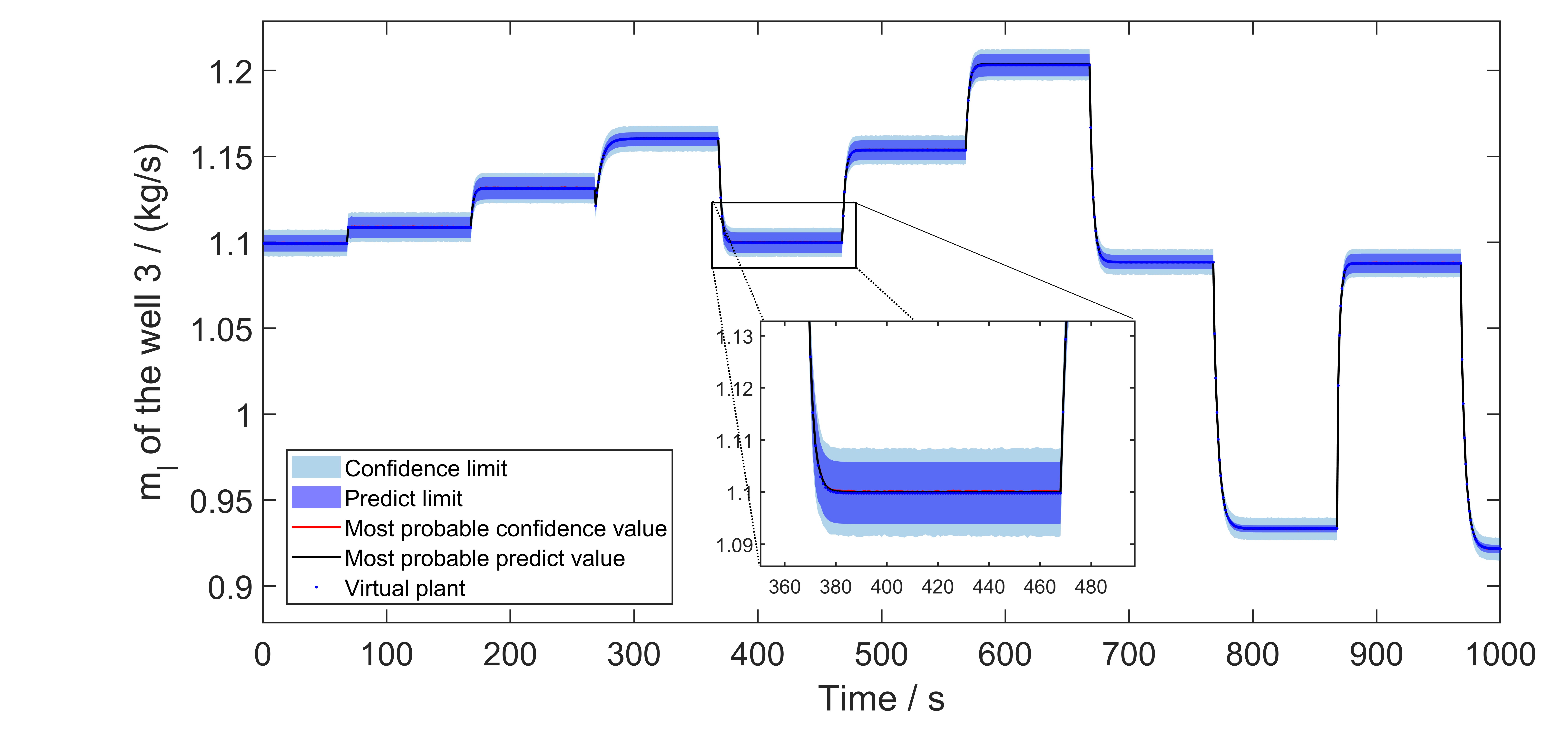}
  \caption{$m_l$ well 3}
  \end{subfigure}
\end{minipage}
\caption{Prediction uncertainty of neural network models sampled for test data}
\label{Figura_parte_2}
\end{figure}

\begin{figure}[h!]
	\centering
		\includegraphics[scale=.05]{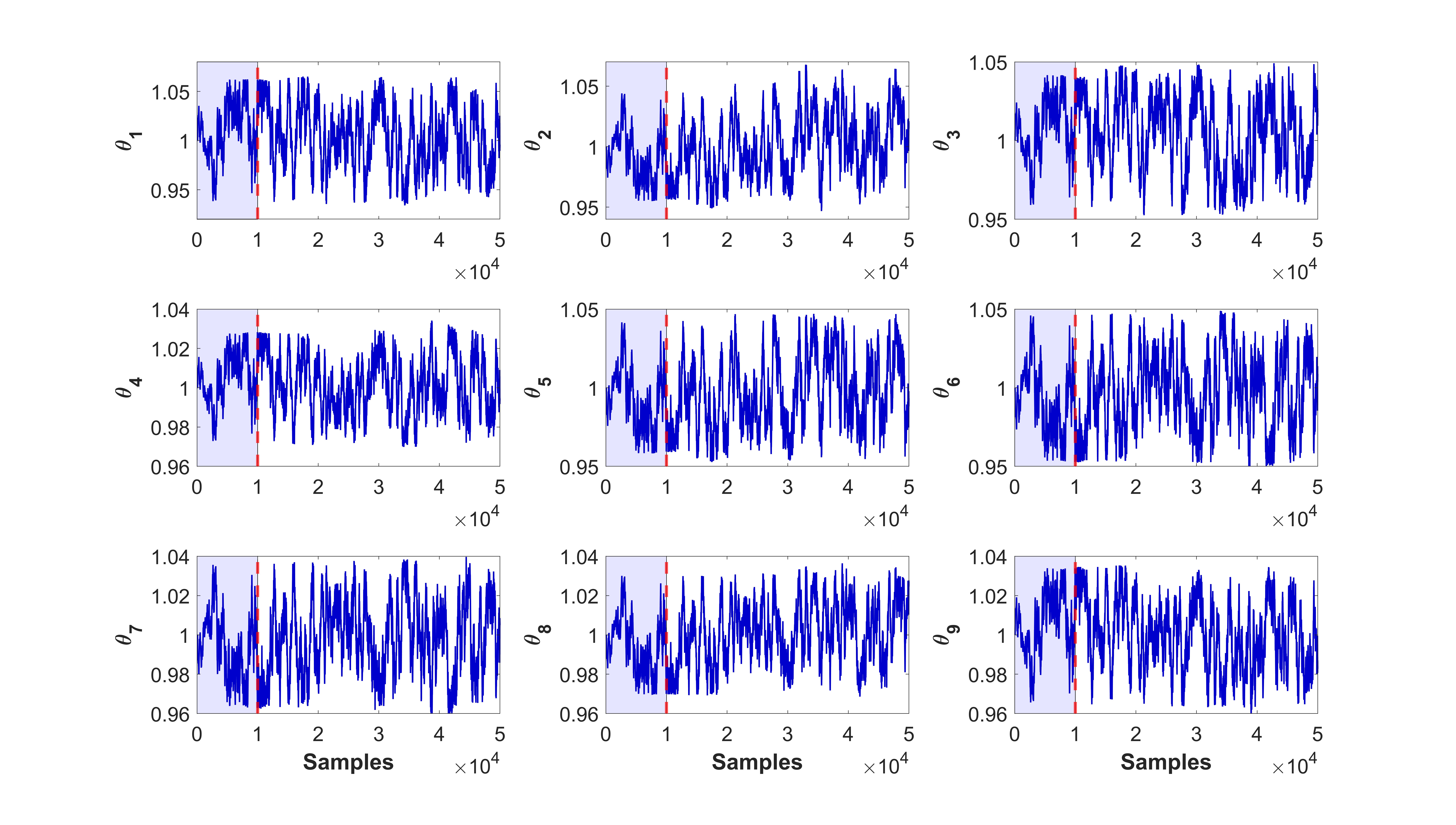}
    \caption{Burn-in MCMC and full chain for nine randomly drawn parameters}
    \label{chain}
\end{figure}

The first step in the uncertainty assessment process is to evaluate the performance of the Markov Chains, which can be evaluated in Figure \ref{chain}. The highlighted area in that figure represents the burn-in phase. This phase is necessary because the work assumes a non-informative prior, which generates low-probability regions that must be removed. The burn-in step minimizes the impact of the first samples on the total samples of the MCMC. This work assumed that the first 10000 samples from the chain correspond to the burn-in phase. The burn-in phase corresponds to the highlighted area in Figure \ref{chain}. Additionally, a total of 50000 samples were used to ensure a comprehensive analysis.

The final step in the proposed methodology is the application of Monte Carlo methods to propagate the identified uncertainty toward the model predictions. This process results in a population of possible models that can represent the system, which together forms the digital twin (DT). By building the confidence region of the DT predictions, we can further evaluate the reliability of the identified DT.

Figure \ref{Figura_parte_1} and Figure \ref{Figura_parte_2} present the DT confidence region plotted against the test data. It is evident from the figure that all test data points are covered by the DT uncertainty, providing a final evaluation of the reliability of the identified DT. This analysis concludes the offline identification step and prepares the DT for deployment.

The use of Monte Carlo methods in this step ensures that the DT is not limited to a single model but instead comprises a range of possible models that can represent the system. This population of models provides a comprehensive representation of the system's behavior, considering the identified uncertainty.

\subsection{Online Digital Twin Implementation}

Finally, the digital twin was deployed in a software-in-the-loop environment. This SIL framework allowed the simulation of a gas lift virtual plant and the deployment of the DT to monitor it. In the present study, we evaluated the performance of the proposed digital twin framework in monitoring a gas lift system through a series of carefully designed scenarios. These scenarios were constructed with the aim of not only testing the robustness and adaptability of our DT model but also exploring its capabilities in different operational conditions. We considered three distinct scenarios for this purpose. Two scenarios where the identification of the drifting source was taken into account. This allowed us to observe how the DT responds to changes in the system over time and how effectively it can identify and adapt to the source of drift. Another scenario that did not consider identifying the drifting source was introduced. This scenario was designed to challenge the DT's adaptability and robustness under unpredictable and potentially disruptive changes. Here, we aimed to investigate how the DT would perform in the face of unexpected deviations in the system's behavior without prior knowledge or detection of the source of drift.

Each of these scenarios provided valuable insights into the performance of our digital twin in monitoring the gas lift system. They also allowed to examine the strengths and limitations of our proposed methodology under different conditions and to identify areas where further improvements may be required. The findings from these scenarios will be discussed in detail in the following sections.

\subsubsection{Scenario 1}
\begin{figure}[h]
\centering
\includegraphics[scale=.032]{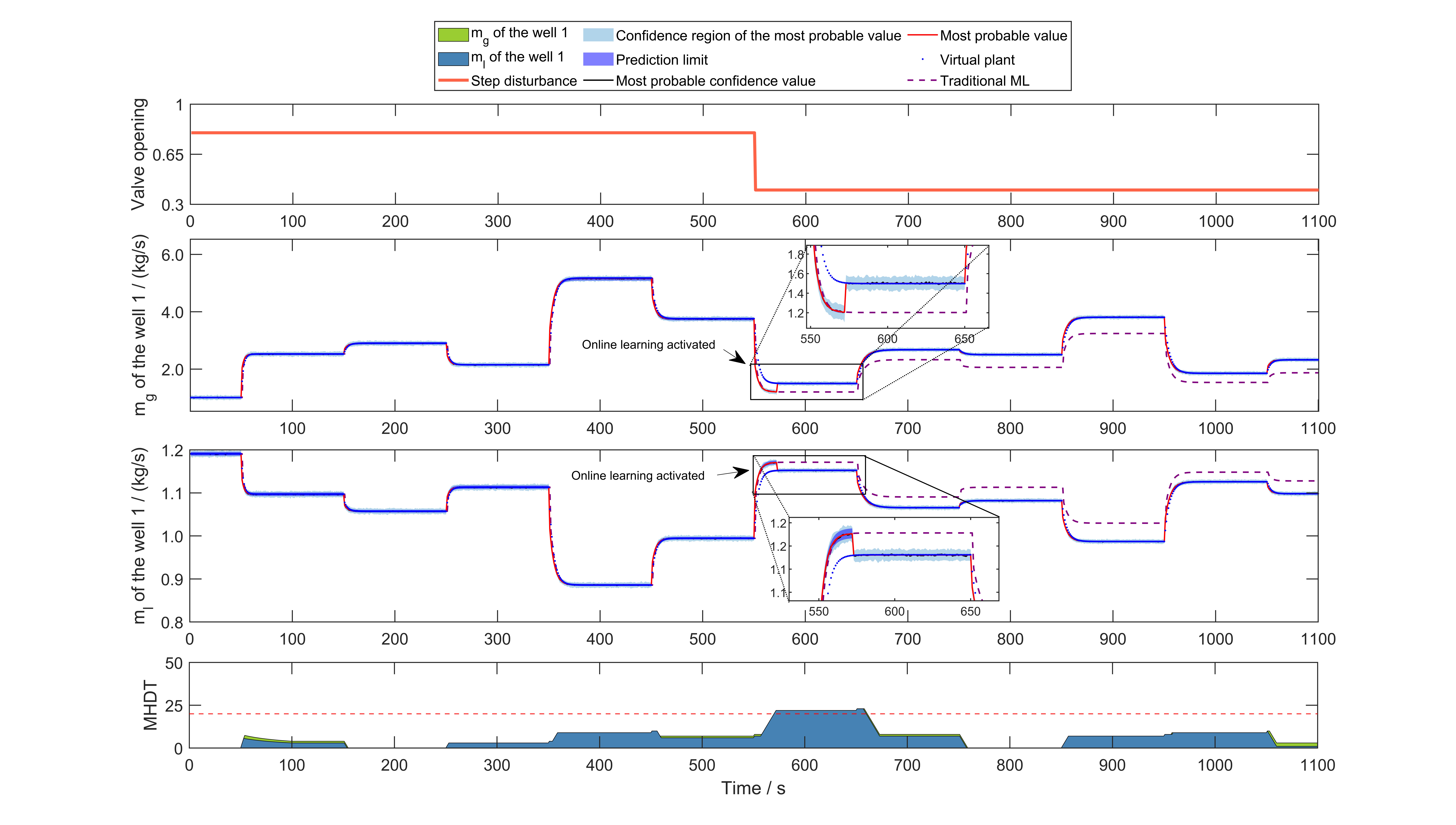}
\caption{Scenario 1 results: The first graph shows a step disturbance in the CV101 valve, the second graph shows the behavior of the variable $m_g$ of the well 1 together with uncertainty evaluation and activation of online learning in the digital twin, third graph the same representation of the previous graph for the variable $ m_l$ of the well 1, and finally the moving horizon for identification of deviation in the digital twin and activation of the cognitive node.}
    \label{cenario_1}
\end{figure}

In Test Scenario 1, the digital twin monitors a gas-lift process and detects degradation in the valve of well 1. The results obtained for this case are presented in Figure \ref{cenario_1}. Unlike scenario 2, the digital twin identifies the source of the problem in this case. Upon recognizing the anomaly, the cognitive node reactivates the offline instance and requests new training data based on the system's current state. Subsequently, data is generated offline using a new design of experiments and sent back to the cognitive node. This cognitive node then activates online learning using these new data to update the predictions of the digital twin.

The results illustrated in Figure \ref{cenario_1} demonstrate that the digital twin experiences a drift during a brief moment, as it is possible to see the zoomed area in Figure \ref{cenario_1}. However, after obtaining the new data and completing the online learning process, the digital twin continues to track the process with remarkable precision. In contrast, the traditional AI model drifts away from the process, unable to adapt to the changing conditions.

The optimal moving horizon was determined by fine-tuning the window size and the sensitivity of the cognitive tracker. By choosing the most suitable window size, the digital twin can strike a balance between collecting adequate data and promptly addressing drifts. Additionally, adjusting the sensitivity helps ensure that the cognitive tracker activates online learning only when needed, avoiding unnecessary system reactions to minor variations, false triggers, or rapid dynamic discrepancies. In this scenario, the optimal cognitive parameters were identified through a sensitivity analysis, resulting in MH=100, and a=1 as the ideal values.

This scenario highlights the benefits of incorporating cognitive capabilities and online learning into digital twin frameworks. The system can maintain high accuracy and reliability throughout the process by enabling the digital twin to recognize drifts and adapt its predictions accordingly. The adaptability and responsiveness demonstrated in this scenario underscore the advantages of the proposed digital twin framework over traditional AI models, which cannot adjust to changing conditions and drifts in real-time.

\subsubsection{Scenario 2}
\begin{figure}[h!]
	\centering
		\includegraphics[scale=.043]{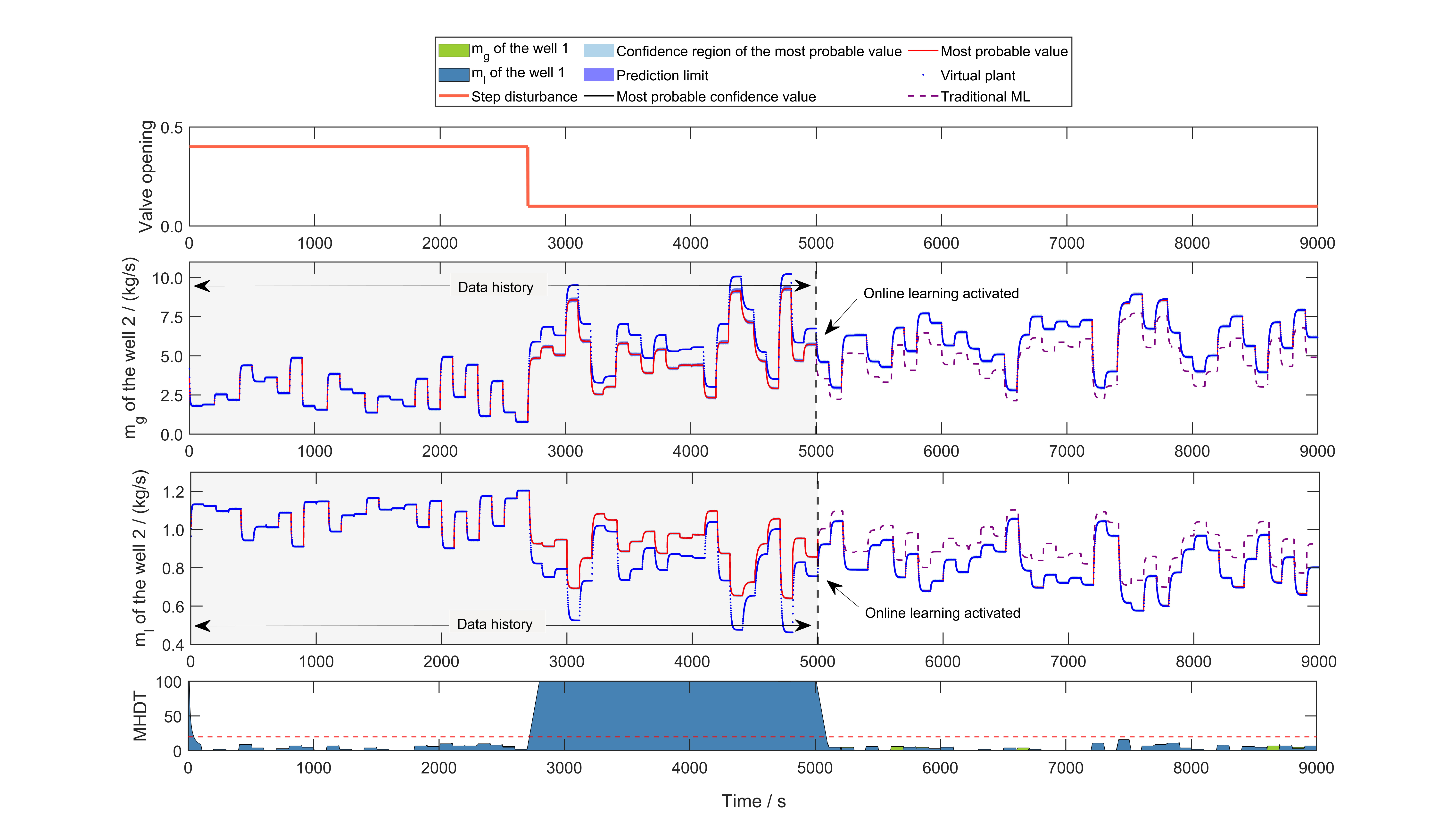}
    \caption{Scenario 2 results: The first graph shows a step disturbance in the CV102 valve, the second graph shows the behavior of the variable $m_g$ of the well 2 along with uncertainty evaluation and activation of online learning in the digital twin, and the third graph shows the exact representation of the previous graph for the variable $ m_l$ of the well 2, and finally the moving horizon for identification of deviation in the digital twin and activation of the cognitive node after a waiting time for data collection.}
    \label{cenario_2}
\end{figure}

In Test Scenario 2, the digital twin monitors a gas-lift process, and at time 2700 s, a reduction of 75\% is observed in the valve CV of well number 2. The digital twin's CT detects a prediction drift, signaling a discrepancy between the virtual model and the actual process. However, the origin of this drift remains unidentifiable for the digital twin, presenting a challenge in determining the appropriate corrective action. This scenario is illustrated in Figure \ref{cenario_2}.

In such situations, the digital twin must wait for sufficient process data to be generated before activating online learning. This waiting period enables the digital twin to collect enough information to identify underlying patterns or trends within the online learning process. During this time, the system relies on its current capabilities to provide information regarding the system. In the present scenario, a waiting period of 5000 seconds is required before the cognitive node activates online learning and corrects the digital twin's prediction. This period is highlighted in Figure \ref{cenario_2}. After this correction, no further drifts are observed for the proposed digital twin. In contrast, Figure \ref{cenario_2} also presents the prediction of a traditional AI model without the enhancements proposed in this work. As evident, the traditional AI model continues drifting, losing its reliability.

To effectively manage this scenario, the moving horizon of the cognitive node, the cognitive tracker, must be properly tuned. If the CT is not calibrated appropriately, the digital twin may activate online learning prematurely or too frequently, leading to unnecessary computational overhead and potentially compromising the system's overall performance.

The moving horizon can be optimized by adjusting the cognitive tracker's window sizes and sensitivity. By selecting an optimal window size, the digital twin can balance the need for sufficient data collection with the urgency of addressing the drift. Fine-tuning the sensitivity ensures that the CT activates online learning only when necessary, preventing the system from overreacting to minor fluctuations, false alarms, or fast dynamic mismatches. In this scenario, the optimal cognitive parameters (MH, and a) were determined through a sensitivity analysis and found to be 100, and 1, respectively. 

Hence, online learning is activated once the digital twin has collected enough process data and the cognitive tracker determines that the drift is significant. The system updates its internal model and uncertainty to better align with the process, improving its predictive accuracy and overall performance. By carefully tuning the moving horizon of the cognitive node, the digital twin can maintain its adaptability and efficiency while effectively addressing the challenges posed by unidentifiable drift origins. However, it is necessary to recognize that this scenario's waiting time is significant. Further studies must be carried out to improve the DT performance in such a situation.

\subsubsection{Scenario 3}

\begin{figure}[h!]
	\centering
		\includegraphics[scale=.032]{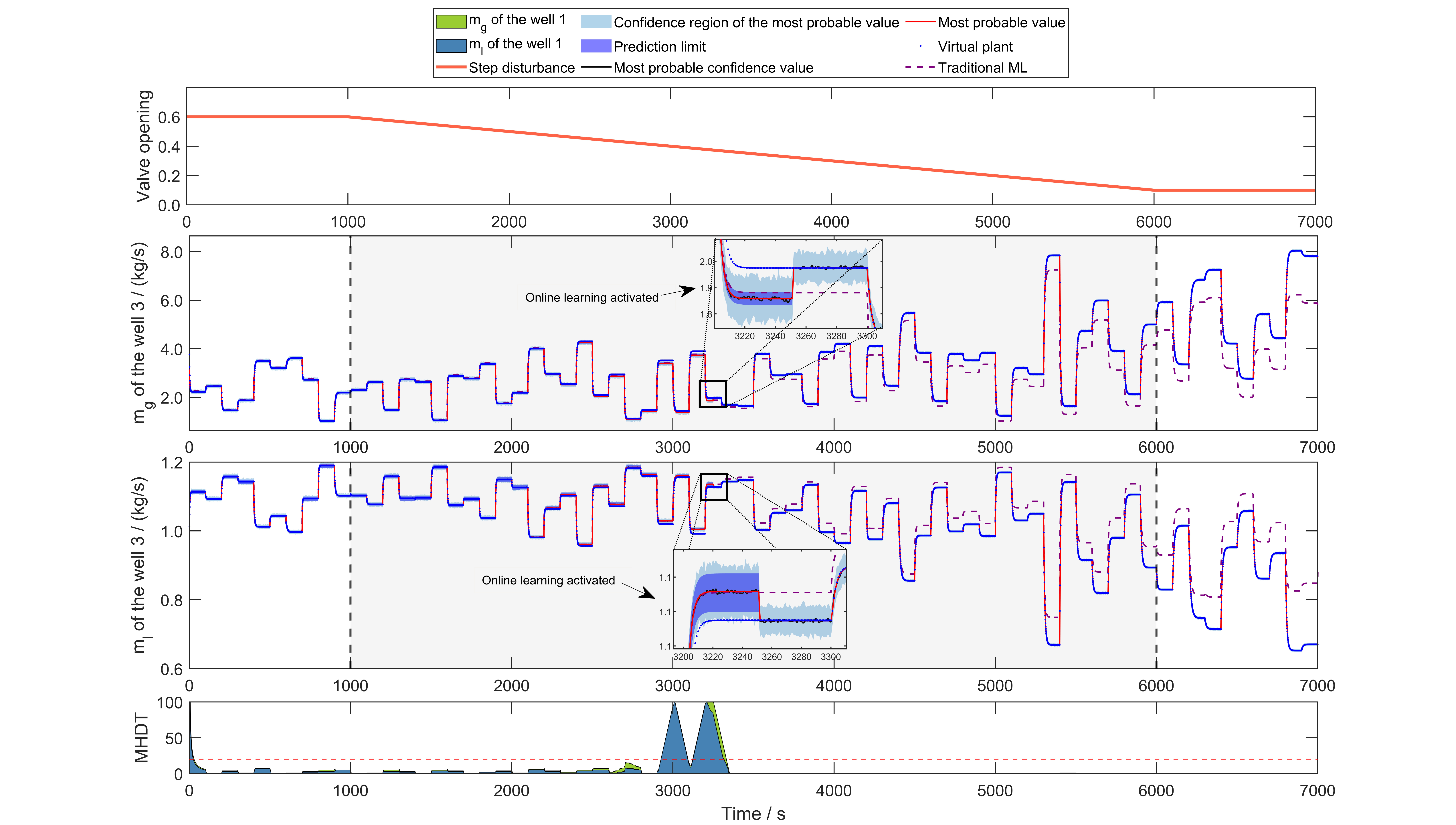}
    \caption{Scenario 3 results: first graph shows a step disturbance in the CV103 valve, second graph shows the behavior of the variable $m_g$ of the well 3 along with uncertainty evaluation and activation of online learning in the digital twin, third graph the same representation of the previous graph for the variable $ m_l$ of the well 3, and finally the moving horizon for identification of deviation in the digital twin and activation of the cognitive node.}
    \label{cenario_3}
\end{figure}

In Test Scenario 3, the digital twin monitors a gas-lift process and observes a continuous degeneration of well number 3. This scenario's results are presented in Figure \ref{cenario_3}. Similar to Scenario 1, the digital twin is able to identify the source of the drifting. Upon recognizing the drift, the cognitive node reactivates the offline instance and requests new training data based on the current state of the system, taking into account the continuous degeneration of well number 3.

Subsequently, data is generated offline using a new design of experiments that accounts for the ongoing degeneration and is sent back to the cognitive node. As the nature of the degeneration depends on the time, the data generation is done for different CV conditions along the drifting slope.  The cognitive node then activates online learning using this new data to update the digital twin's predictions, allowing it to adapt to the changing conditions in the well.

As the degeneration continues, the digital twin effectively tracks the process and maintains high accuracy and reliability. Its adaptability and responsiveness to the continuous degeneration of well number 3 demonstrate the benefits of incorporating cognitive capabilities and online learning into the digital twin framework. By recognizing drifts and updating its predictions accordingly, the system is better equipped to handle dynamic changes in the process and maintain its performance, in contrast to traditional AI models that lack the ability to adjust in real time.

\section{Conclusions}
This study presented an innovative and comprehensive digital twin framework designed to facilitate optimal and autonomous decision-making, specifically focusing on gas-lift processes in the oil and gas industry. This framework amalgamates several key techniques, including offline training, Bayesian inference, Monte Carlo simulations, transfer learning, online learning, and a novel model hyperspace dimension reduction with cognitive tack. Integrating these techniques results in an adaptive, robust, and efficient system while addressing the computational challenges associated with online learning and the accurate representation of uncertainty in AI-based systems.
Developing AI models that balance computational efficiency with robustness and adaptability is crucial for industries like oil and gas, where operations must be carried out under strict economic, safety, and environmental constraints. However, this is not only limited to the oil and gas industries. By leveraging the proposed digital twin framework, industrial operations can enhance their operations safety, reliability, and precision. Integrating AI models and digital twins enables better monitoring and management of complex processes, ultimately reducing the risks associated with uncertainty and unforeseen events. The fusion of digital twin technology with AI has the potential to optimize industrial processes, minimize environmental impacts, and increase overall operational efficiency.
The results of this study contribute to the expanding knowledge base surrounding digital twin technology and its applications in the oil and gas industry. The proposed framework establishes a foundation for future research, especially in robust and adaptive AI models for complex systems. While the current methodology offers numerous advantages, there are opportunities to refine and expand upon certain aspects to address the limitations discussed in the results section.
For instance, although the digital twin can effectively track and maintain up-to-date information, it experiences considerable delays when the source of prediction drifting is unidentifiable. To overcome this limitation, future research could explore integrating fault detection tools into the digital twin framework. This enhancement would enable the system to identify and respond to discrepancies better, improving its adaptability and overall performance.

In summary, this study not only adds to the growing literature on digital twin technology in the oil and gas industry but also provides a foundation for developing more robust and adaptive AI models. 
In summary, this study demonstrates the potential of a comprehensive digital twin framework to address the challenges associated with optimal and autonomous decision-making. Industries can improve safety, reliability, and precision in their operations by harnessing the power of AI models that are computationally efficient, robust, and adaptive.

\bibliographystyle{model1-num-names}

\bibliography{cas-refs}





\end{document}